\documentclass[sigconf]{acmart}
\AtBeginDocument{%
  \providecommand\BibTeX{{%
    \normalfont B\kern-0.5em{\scshape i\kern-0.25em b}\kern-0.8em\TeX}}}

\usepackage{booktabs}
\usepackage{multicol}
\usepackage{multirow}
\usepackage{graphicx}
\usepackage{subcaption}

\copyrightyear{2022} 
\acmYear{2022} 
\setcopyright{acmcopyright}
\acmConference[CIKM '22]{Proceedings of the 31st ACM International Conference on Information and Knowledge Management}{October 17--21, 2022}{Atlanta, GA, USA}
\acmBooktitle{Proceedings of the 31st ACM International Conference on Information and Knowledge Management (CIKM '22), October 17--21, 2022, Atlanta, GA, USA}
\acmPrice{15.00}
\acmDOI{10.1145/3511808.3557091}
\acmISBN{978-1-4503-9236-5/22/10}

\begin{document}

\title[DuETA: Traffic Congestion Propagation Pattern Modeling via \\Efficient Graph Learning for ETA Prediction at Baidu Maps]{DuETA: Traffic Congestion Propagation Pattern Modeling via Efficient Graph Learning for ETA Prediction at Baidu Maps}

\author{Jizhou Huang}
\authornote{These authors contributed equally to this work.}
\authornote{Corresponding author: Jizhou Huang.}
\email{huangjizhou01@baidu.com}
\orcid{0000-0003-1022-0309}

\author{Zhengjie Huang}
\authornotemark[1]
\email{huangzhengjie@baidu.com}
\orcid{0000-0003-1878-0554}

\author{Xiaomin Fang}
\authornotemark[1]
\email{fangxiaomin01@baidu.com}
\orcid{0000-0002-7563-5268}

\affiliation{%
  \institution{Baidu Inc.}
  \streetaddress{Shangdi 10th Street}
  \city{Haidian District}
  \state{Beijing}
  \country{China}}
  
\author{Shikun Feng}
\authornotemark[1]
\email{fengshikun01@baidu.com}
\orcid{0000-0002-0191-4854}

\author{Xuyi Chen}
\email{chenxuyi@baidu.com}
\orcid{0000-0003-1368-2277}

\affiliation{%
  \institution{Baidu Inc.}
  \streetaddress{Shangdi 10th Street}
  \city{Haidian District}
  \state{Beijing}
  \country{China}}

\author{Jiaxiang Liu}
\email{liujiaxiang@baidu.com}
\orcid{0000-0002-6369-1609}

\author{Haitao Yuan}
\email{yuanhaitao@baidu.com}
\orcid{0000-0003-2001-8614}

\author{Haifeng Wang}
\email{wanghaifeng@baidu.com}
\orcid{0000-0002-0672-7468}

\affiliation{%
  \institution{Baidu Inc.}
  \streetaddress{Shangdi 10th Street}
  \city{Haidian District}
  \state{Beijing}
  \country{China}}

\renewcommand{\shortauthors}{Jizhou Huang et al.}

\begin{abstract}
Estimated time of arrival (ETA) prediction, also known as travel time estimation, is a fundamental task for a wide range of intelligent transportation applications, such as navigation, route planning, and ride-hailing services.
To accurately predict the travel time of a route, it is essential to take into account both contextual and predictive factors, such as spatial-temporal interaction, driving behavior, and traffic congestion propagation inference.
The ETA prediction models previously deployed at Baidu Maps have addressed the factors of spatial-temporal interaction (ConSTGAT \cite{fang2020constgat}) and driving behavior (SSML \cite{fang2021ssml}).
In this work, we believe that modeling traffic congestion propagation patterns is of great importance toward accurately performing ETA prediction, and we focus on this factor to improve ETA performance.
Traffic congestion propagation pattern modeling is challenging, and it requires accounting for impact regions over time and cumulative effect of delay variations over time caused by traffic events on the road network.
In this paper, we present a practical industrial-grade ETA prediction framework named DuETA.
Specifically, we construct a congestion-sensitive graph based on the correlations of traffic patterns, and we develop a route-aware graph transformer to directly learn the long-distance correlations of the road segments.
This design enables DuETA to capture the interactions between the road segment pairs that are spatially distant but highly correlated with traffic conditions. 
Extensive experiments are conducted on large-scale, real-world datasets collected from Baidu Maps.
Experimental results show that ETA prediction can significantly benefit from the learned traffic congestion propagation patterns, which demonstrates the effectiveness and practical applicability of DuETA.
In addition, DuETA has already been deployed in production at Baidu Maps, serving billions of requests every day. This demonstrates that DuETA is an industrial-grade and robust solution for large-scale ETA prediction services.
\end{abstract}



\maketitle

\section{Introduction}

\begin{figure}[t!]
\centering
\includegraphics[width=1.0\columnwidth,trim={0cm 0cm 0cm 0.9cm},clip]{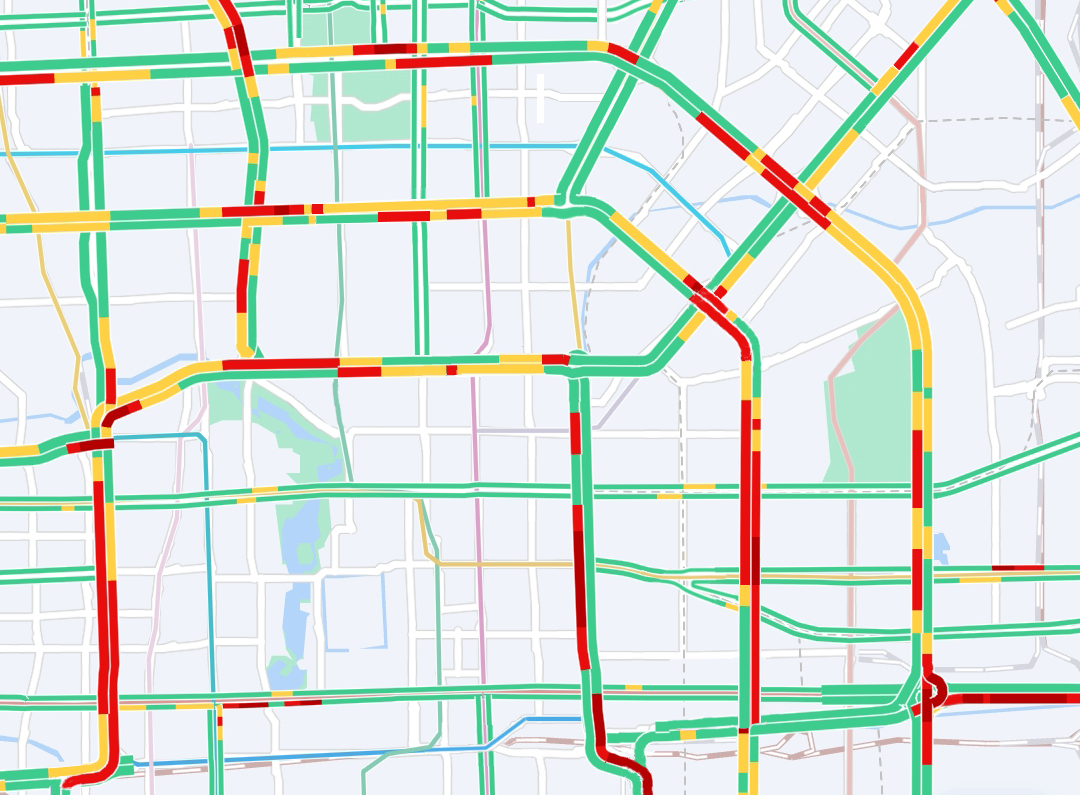}
\vspace{-5mm}
\caption{A screenshot of live traffic service at Baidu Maps.}
\label{fig:ltce}
\vspace{-3mm}
\end{figure}

Estimated time of arrival (ETA) prediction (a.k.a., travel time estimation) aims at predicting the travel time for a given route and departure time, which greatly helps users to make informed decisions about traffic conditions and plan their travels wisely in advance.
ETA prediction is a fundamental task for a wide range of intelligent transportation applications, such as navigation, route planning, and ride-hailing services.
As one of the largest web mapping applications, Baidu Maps keeps serving tens of billions of daily ETA requests that benefit tens of millions of users per day.
In order to help traffic participants make more informed decisions on route selection and congestion avoidance, it is important to provide accurate and reliable travel time estimations.

\begin{figure}[!t]
\setlength{\abovecaptionskip}{0.01cm} 
\setlength{\belowcaptionskip}{0.01cm}
    \centering
    \includegraphics[width=.79\columnwidth,trim={0.0cm 0.21cm 0.0cm 0.0cm},clip]{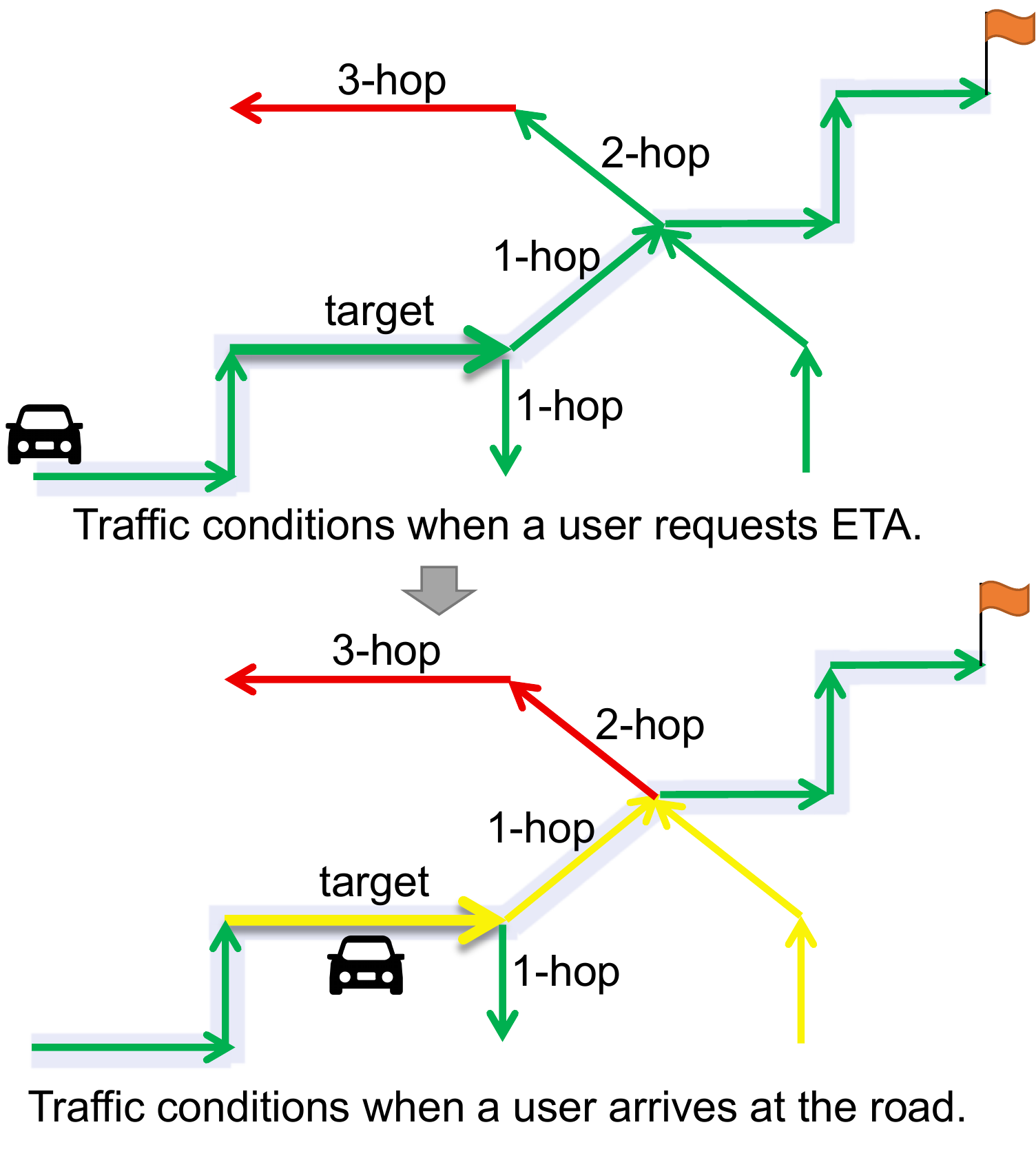}
    \caption{A representative case of associations between the road segments that are not directly connected. An arrow indicates a road segment. The road segments with gray background constitute the route selected by a user. The live traffic condition of each road segment is marked with a distinctive color, i.e., ``congestion''/``slow''/``fast'' are in red/yellow/green, respectively.}
    \label{fig:cases}
\vspace{-4mm}  
\end{figure}

ETA prediction is a challenging task, as it needs to take into account both contextual and predictive factors, such as spatial-temporal interaction, driving behavior, and traffic congestion propagation inference.
The ETA prediction models previously deployed at Baidu Maps have addressed the factors of spatial-temporal interaction (ConSTGAT \cite{fang2020constgat}) and driving behavior (SSML \cite{fang2021ssml}).
In our efforts toward developing a more powerful ETA prediction model, we observed that a propagation of ETA errors arises from the sharp inconsistency between the predicted traffic condition in the future and ground truth.
As such, besides the current traffic conditions on the road network, it is also important to accurately infer the traffic conditions unfolding in the future.
Motivated by this observation, we focus on modeling traffic congestion propagation patterns to improve ETA performance in this work.
Figure \ref{fig:ltce} shows an example of live traffic conditions at Baidu Maps.
As illustrated in it, the impact regions and cumulative delays over time caused by traffic congestion (the road segments in red) would inevitably affect all the interdependent segments on the road network.

Although there is a growing interest in incorporating traffic conditions into ETA prediction models, e.g., \cite{zhang2018gaan,li2018diffusion,yu2018spatio-temporal,guo2019attention,yu20193d,fang2020constgat}, the inferring of traffic conditions unfolding in the future still remains a bottleneck in industrial ETA prediction models, particularly for modeling impact regions and cumulative delays over time caused by traffic congestion.
To illustrate the importance of modeling traffic congestion propagation patterns in ETA prediction models, we consider the representative case presented in Figure \ref{fig:cases}.

As can be seen from Figure~\ref{fig:cases}, road segment pairs that are spatially distant and indirectly connected can interact with traffic conditions, which demonstrates the importance of modeling impact regions over time caused by traffic congestion.
Although existing studies have applied spatial-temporal graph neural networks (STGNNs) \cite{zhang2018gaan,li2018diffusion,yu2018spatio-temporal,guo2019attention,yu20193d,fang2020constgat} to model traffic conditions, they pay too much attention to directly connected road segments, which have two main limitations. (1) The long-distance correlations of indirectly connected road segments are not explicitly modeled, which inevitably suffer from information loss during the multi-step message passing. (2) Traffic conditions are not sufficiently transmitted between two road segments that are spatially distant, because they typically execute only a few steps of message passing (one step in most cases), due to the computational complexity of STGNNs.

In this paper, we present our efforts toward designing and developing DuETA, which is designed to model traffic congestion propagation patterns via efficient graph learning.
Specifically, instead of directly using the road network as a graph, we construct a congestion-sensitive graph based on the correlations of traffic patterns.
In addition, we develop a route-aware graph transformer to directly learn the long-distance correlations of the road segments. 
These designs enable DuETA to capture the interactions between any two road segment pairs that are spatially distant but highly correlated with traffic conditions. 

Extensive experiments are conducted on large-scale, real-world datasets collected from Baidu Maps.
Experimental results show that DuETA is able to effectively learn traffic congestion patterns and predict subsequent propagation of traffic events, which significantly benefits the performance of ETA prediction.

Our main contributions to this problem are as follows:
\begin{itemize}
\item \textbf{Potential impact:} We suggest a practical and robust framework, named DuETA, as an industrial-grade solution to the task of ETA prediction. We hope that it could be of potential interest to practitioners working with such problems. 
\item \textbf{Novelty:} The design of DuETA is driven by the novel ideas that directly capture the long-distance correlations through a congestion-sensitive graph, and that model traffic congestion propagation patterns via a route-aware graph transformer.
\item \textbf{Technical quality}: Extensive experiments show that ETA prediction can significantly benefit from the learned traffic congestion propagation patterns, which demonstrates the effectiveness and practical applicability of DuETA. The successful deployment of DuETA at Baidu Maps further shows that it is an industrial-grade and robust solution for large-scale ETA prediction services.
\end{itemize}

\section{D\texorpdfstring{\MakeLowercase{u}}{u}ETA}
We first formalize the task, then detail the architecture of DuETA. 

\subsection{Problem Formulation}
When the ETA prediction service receives a request, it will provide the estimated arrival time on the basis of the received request, road network, road conditions, and other contextual information. 

\textbf{Road network:} The road network is an essential component of ETA. In this study, the road network is defined as a directed graph $\mathcal{G}=(\mathcal{L},\mathcal{E})$, where $\mathcal{L}$ is a link set and $\mathcal{E}$ is an edge set. Link $l \in \mathcal{L}$ represents a road segment. For the sake of convenience, ``road segment'' is referred to as ``link'' hereafter. Edge $e_{ij} \in \mathcal{E}$ denotes the edge connecting link $l_{i}$ and link $l_{j}$, if ${l}_i$ and ${l}_j$ share the same junction.

\textbf{Route}: A route $r$ is defined as a link sequence $r=[l_1, l_2, \cdots, l_m]$, where $m$ is the number of links in the route. Usually, a route contains hundreds of consecutive links. A navigation service produces several candidate routes based on the corresponding road network.

\textbf{Request}: A request is represented by a pair $req = (r,s)$, where $r$ is the route, and $s$ is the departure time. The objective of ETA is to estimate the travel time $y$ of the given request $req$.

\textbf{Dataset}: A dataset is defined as $\mathcal{D}=\{(req^{(i)}, y^{(i)})\}_{i=1}^{n}$, where $y^{(i)}$ is the ground-truth travel time of $r$ in $req^{(i)}$, and $n$ is the number of requests in the dataset. For $(req^{(i)}, y^{(i)}) \in \mathcal{D}$, the travel time of the $j$-th link $l_{j}^{(i)}$ in $r^{(i)}$ of request $req^{(i)}$ is denoted as $y_{j}^{(i)}$, and the travel time of the entire route is computed by $y^{(i)}=\sum_{j=1}^{m^{(i)}}y_{j}^{(i)}$.

\subsection{Feature Preparation}
Two types of features are prepared for ETA prediction: static features and dynamic features. The static features refer to features that do not change over time, consisting of the information of the road network of the corresponding city (e.g., the link ID, the length, the width, the number of lanes, the type of road, the speed limit, the type of crossing, and the kind of traffic light), as well as the contextual information (e.g., departure time and user profile).

On the contrary, the dynamic features change over time, such as traffic conditions. We aim to infer the future traffic conditions from the recent traffic conditions. When estimating the time of arrival, the traffic conditions of the past one hour are collected as features, which are divided into 12 time slots (5 minutes per time slot). For each link at each time slot, we calculate the median speed, max speed, min speed, mean speed, and record counts as features. These statistic numbers of a link $l$ at time slot $t$  are deduced from the traffic records in the dataset, where a time slot is set to 5 minutes in this paper. Then, these features are flattened and mapped into the same shape with static features by a linear transformation. Static and dynamic features are taken as the features of the links. For link $l_i \in \mathcal{L}$, we denote its feature vector as $\mathbf{x}_i$.

\subsection{Congestion-sensitive Graph}
Instead of directly using the road network $\mathcal{G}$ as the graph for traffic congestion pattern modeling, we design a congestion-sensitive graph $\mathcal{G}^{cs}=(\mathcal{L}, \{\mathcal{E}_r^{f}\}|_{r=1}^5, \mathcal{E}^{h})$, where $\mathcal{L}$ is the link set. For each link $l \in \mathcal{L}$, we take advantage of the first-order neighbor links, as well as the high-order neighbor links whose traffic patterns are highly correlated to that of link $l$. $\{\mathcal{E}_r^{f}\}$ is a set of edge sets that describes the first-order neighbors, while $\mathcal{E}^{h}$ is an edge set that describes the high-order neighbors. We detail them as follows.

\subsubsection{First-order Neighbors}
An edge $e_{ij} \in \mathcal{E}$ of the road network $\mathcal{G}$ describes the relation between a link $l_i$ and its first-order neighbor $l_j$ that is directly connected to $l_i$.
Our previous work \cite{fang2020constgat} has demonstrated that different types of neighbor links play varying roles in the future traffic of a given link. For example, the traffic congestion is more likely to propagate from downstream links to upstream links. 
In order to effectively handle such correlations, the following refinements are performed.
First, we define multiple types of link relations and incorporate these relations into the construction of the congestion-sensitive graph. Second, we use attention mechanism separately for each relation to capture the impact of neighbor links, which is detailed in Section~\ref{sec:base_gt} and \ref{sec:route-aw}.

\begin{figure}[!t]
\setlength{\abovecaptionskip}{0.16cm} 
    \centering
    \includegraphics[width=1.0\linewidth]{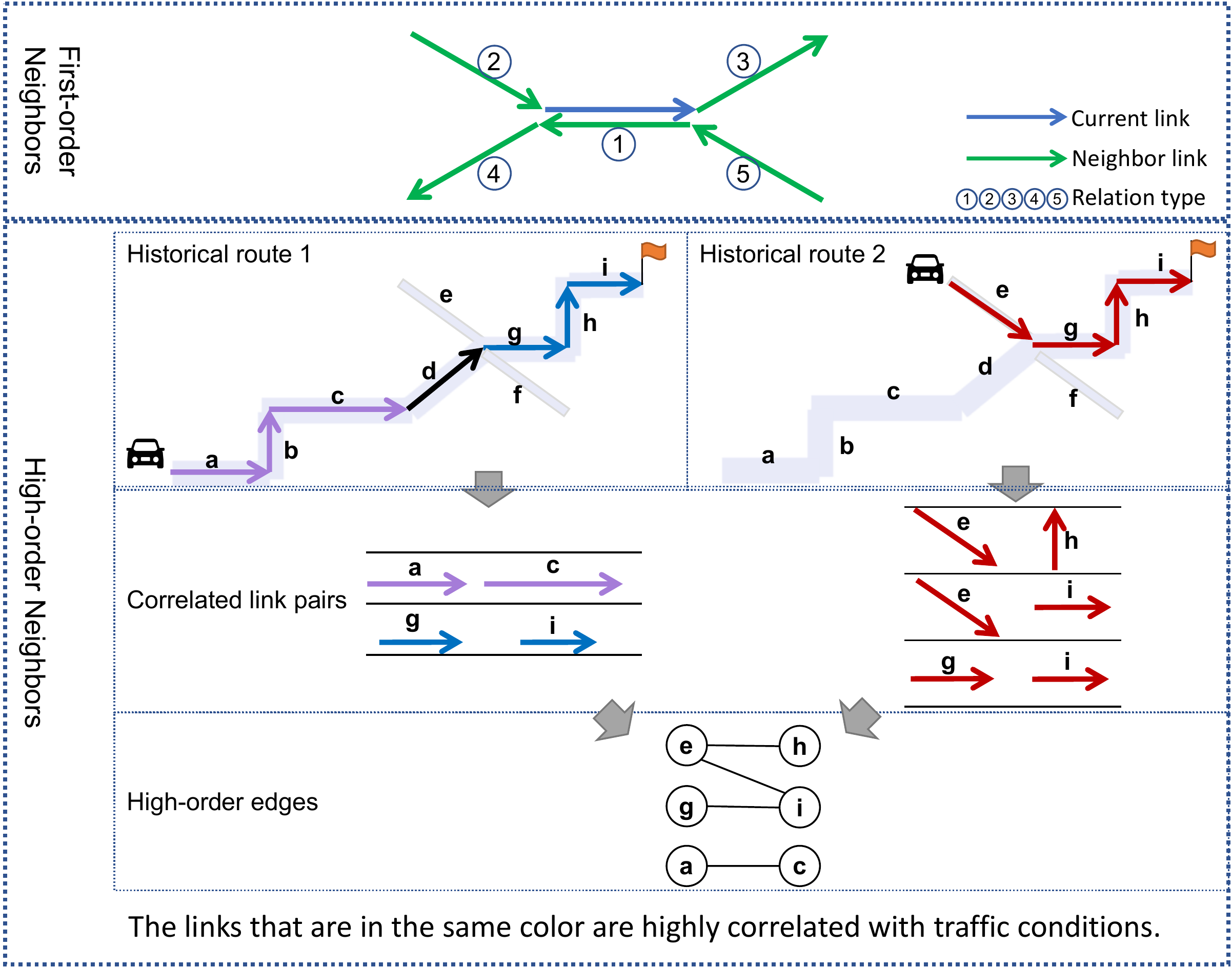}
    \caption{Demonstration of extracting first-order and high-order neighbors to construct a congestion-sensitive graph.}
    \label{fig:congestion_sensitive_graph}
\vspace{-3.5mm}
\end{figure}

An edge describes the relation between two links, and all the edges in the edge set $\mathcal{E}$ are divided into five types according to the connection relationships between the links. For example, as shown on the top of Figure~\ref{fig:congestion_sensitive_graph}, given a link, the second type of the first-order neighbor is its upstream link, and the third type of the first-order neighbor is its downstream link. Although the links of the remaining three types are not included in the travel route, the traffic conditions of these links also affect that of the target link. For example, the vehicles on these links may block the traffic at the intersection, thus holding up the traffic of the target link. The set for each type of edge is denoted by $\mathcal{E}^{f}_r$ with $r$ denoting the index of the type. That means $\mathcal{E}=\cup\{\mathcal{E}^{f}_r\}|_{r=1}^5$. 

\begin{figure*}[htbp]
\setlength{\abovecaptionskip}{0.15cm} 
\centering
\includegraphics[width=0.95\linewidth]{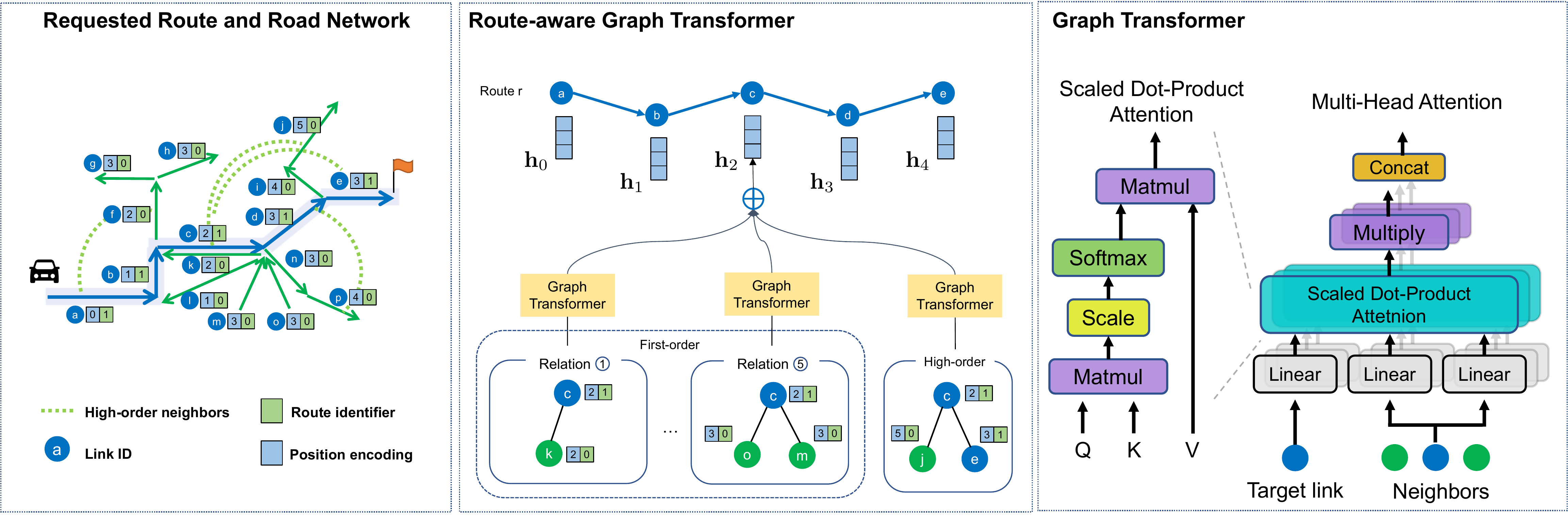}
\caption{Architecture of the route-aware graph transformer.}
\label{fig:overall_archi}
\vspace{-5mm}
\end{figure*}

\subsubsection{High-order Neighbors}
\label{sec:high_corr}
Existing studies only focus on the relationships between directly connected road segments. Despite the associations between the directly connected links, the long-distance associations between indirectly connected links are also crucial for ETA prediction. 
We aim to model the interactions between the link pairs that are spatially distant but highly correlated. Given a link, its indirectly connected links are called the high-order neighbors. Theoretically, all the indirectly connected links in the road network can be regarded as its high-order neighbors, and there are a great number of links in the road network. Considering the computational cost of STGNN, we extract the most correlated high-order neighbors for each link to construct the congestion-sensitive graph.

The correlated high-order neighbors are extracted from tens of millions of historical travel routes of Baidu Maps.
The idea of high-order neighbor construction is demonstrated in Figure \ref{fig:congestion_sensitive_graph}.
We pay attention to each link's 2-hop to 5-hop neighborhood to extract the correlated high-order neighbors, since traffic correlation between two links too far away is usually low. More concretely, for a link $l_i$ in a historical travel route $r=[l_1, l_2, \cdots, l_m]$, we first extract its 2-hop to 5-hop neighbors in the route as the candidate  high-order neighbors of link $l_i$. 
We denote those candidate high-order neighbors w.r.t. route $r$ as $N_{2-5}^r(l_i)=\{l_{i-5}, l_{i-4}, l_{i-3}, l_{i-2}, l_{i+2}, l_{i+3}, l_{i+4}, l_{i+5}\}$. Then, we calculate the Pearson correlation\footnote{To calculate the Pearson correlation between two links $l_1$ and $l_2$, we first count the average travel time every five minutes for the last two hours as $T_1=[t_1^0, t_1^1, \cdots, t_1^{23}]$ and $T_2=[t_2^0, t_2^1, \cdots, t_2^{23}]$. Then, the Pearson correlation is computed by $\frac{\text{cov}(T_1, T_2)}{\rho_{T_1}\/\rho_{T_2}}$.} between the traffic patterns of link $l_i$ and each candidate in $N_{2-5}^r(l_i)$. The route-based correlation score between two links in the historical travel route $r$ is denoted as $c_{ij}^r$. We make the assumption that the higher the correlation score, the higher the probability that the corresponding links will impact each other. Finally, accounting for the population of co-occurrence of two links, for links $l_i$ and $l_j$, we sum up all the route-based correlation scores $c_{ij}^r$ of all the historical travel routes as the final correlation score $c_{ij}^{final}$. The most influential high-order neighbors of link $l_i$ are defined as those links with the highest correlation scores $c_{ij}^{final}$. For each link, we select the top-5 influential high-order neighbors. 
A high-order edge is defined as an edge that connects a link and one of its high-order neighbor links. All the high-order edges compose a high-order edge set, i.e., $\mathcal{E}^h$. Note that $\mathcal{E}^h$ is regarded as a supplemental edge set to capture the associations between the links that are spatially distant but highly correlated, and $\mathcal{E} \cap \mathcal{E}^h=\emptyset$.

For the sake of convenience, we will use $\mathcal{G}^{CS}=(\mathcal{L},\{\mathcal{E}_r\}|_{r=1}^{6})$ to denote the congestion-sensitive graph $\mathcal{G}^{CS}=(\mathcal{L}, \{\mathcal{E}_r^f\}|_{r=1}^5, \mathcal{E}^h)$ hereafter, where the high-order edges are considered as the sixth type of link relations.

\subsection{Route-aware Graph Transformer}
We design a route-aware graph transformer, which aims at efficiently and effectively aggregating traffic information through the congestion-sensitive graph. Figure \ref{fig:overall_archi} illustrates the architecture of the route-aware graph transformer. It takes the congestion-sensitive graph and the prepared features as input and generates the representation vector of each link in the route. 

\subsubsection{Graph Transformer}
\label{sec:base_gt}
In order to specify different weights to different links in a neighborhood, we use the graph transformer \cite{shi2020masked} as the backbone structure to aggregate the information from our congestion-sensitive graph. The reasons for choosing the graph transformer are two-fold. (1) It adopts the multi-head attention mechanism \cite{vaswani2017attention} to learn edge weights. (2) It addresses the over-smoothing problem in vanilla GNNs by residual connections.

Given a graph $\mathcal{G}=(\mathcal{L}, \mathcal{E})$ and the corresponding link features $\mathbf{X}=[\mathbf{x}_1, \mathbf{x}_2, \cdots, \mathbf{x}_N]$ of route $r=[l_1, l_2, \cdots, l_N]$, the graph transformer performs multi-head attention for each edge $e_{ij} \in \mathcal{E}$:
\begin{equation}
\mathbf{q}_{c,i}=\mathbf{W}_{c}^Q\mathbf{x}_i + \mathbf{b}_{c}^Q,
\end{equation}
\begin{equation}
\mathbf{k}_{c,j}=\mathbf{W}_{c}^K\mathbf{x}_j + \mathbf{b}_{c}^K,
\end{equation}
\begin{equation}
\mathbf{v}_{c,j}=\mathbf{W}_{c}^V\mathbf{x}_j + \mathbf{b}_{c}^V,
\end{equation}
\begin{equation}
\alpha_{c,i,j}=\frac{\langle \mathbf{q}_{c,i}, \mathbf{k}_{c,j} \rangle }{\sum_{k\in \mathcal{N}(i)} \langle \mathbf{q}_{c,i}, \mathbf{k}_{c,k} \rangle },
\end{equation}
where $q_{c,i}$, $k_{c,j}$, and $v_{c,j}$ denote the query, key, and value, respectively, of the attention mechanism. $c$ represents the index of the head. For the $c$-th head attention, we transform the feature vector of link $l_i$, i.e., $\mathbf{x}_i$ into a query vector $\mathbf{q}_{c,i} \in \mathbb{R}^{d}$. The feature vector of link $l_j$ (a neighbor of link $l_i$), i.e., $\mathbf{x}_j$ is converted into a key vector $\mathbf{k}_{c,j}\in\mathbb{R}^{d}$ and a value vector $\mathbf{v}_{c,j}\in\mathbb{R}^{d}$. $\mathbf{W}_{c}^Q$, $\mathbf{b}_c^Q$, $\mathbf{W}_c^K$, $\mathbf{b}_c^K$, $\mathbf{W}_c^V$, and $\mathbf{b}_c^V$ are learnable parameters.
$\langle \mathbf{q}, \mathbf{k} \rangle =\exp(\frac{\mathbf{q}^{T}\mathbf{k}}{\sqrt{d}})$ is an exponential scale dot-product function, and $d$ is the hidden size of each head. $N(i)$ denotes the neighbor links of link $l_i$ according to the edge set $\mathcal{E}$. With the graph multi-head attention $\alpha_{c,i,j}$, we perform weighted aggregation from all the neighbors of link $i$:
\begin{equation}
	\mathbf{h}_i= \mathbf{x}_i + \frac{1}{C}\sum_{c=1}^{C} \sum_{j\in \mathcal{N}(i)} \alpha_{c,i,j}\mathbf{v}_{c,j},
	\label{equ:graph_transformer}
\end{equation}
where $C$ is the number of attention heads. To tackle the over-smoothing issue \cite{chen2020simple, li2019deepgcns}, we simply average the results from different attention heads to keep the same shape with the residual connection term $\mathbf{x}_i$. 
For the sake of convenience, function $\mathbf{h}_i=\text{GT}(i, \mathcal{G}, \mathbf{X})$ is introduced to denote the graph transformer.

\subsubsection{Route-aware Graph Transformer}
\label{sec:route-aw}
In order to better capture the correlations between different links, we decompose the graph $\mathcal{G}^{CS}=(\mathcal{L}, \{\mathcal{E}_r\}|_{r=1}^6)$ into six relational graphs, i.e., $\mathcal{G}^{(r)}=(\mathcal{L}, \mathcal{E}_r)$ where the relation index $r$ is from 1 to 6. We apply graph transformer to each relational graph $\mathcal{G}^{(r)}$ to obtain the representations of links. Then, for each link $l_i$, we summarize its representations that are learned through different relational graphs to produce a new hidden vector, which incorporates the six kinds of relations. Equation~\ref{equ:graph_transformer} is rewritten as: 
\begin{equation}
	\label{equ:relation_aggr}
 	\mathbf{h}_{i}= \mathbf{x}_i + \frac{1}{6C }\sum_{r=1}^6\sum_{c=1}^{C} \sum_{j\in \mathcal{N}_{r}(i)} \alpha_{c,i,j}^{(r)}\mathbf{v}^{(r)}_{c,j},
\end{equation}
where $\mathcal{N}^{(r)}(i)$ is the set of neighbors for link $i$ w.r.t. the $r$-th relation.

\begin{figure}[!th]
\setlength{\abovecaptionskip}{0.1cm}
	\centering
	\begin{subfigure}{1.0\columnwidth}
	    \setlength{\abovecaptionskip}{0.1cm}
        \setlength{\belowcaptionskip}{0.01cm}
		\centering
		\includegraphics[width=1.0\columnwidth,trim={0.0cm 0.5cm 0.0cm 0.0cm},clip]{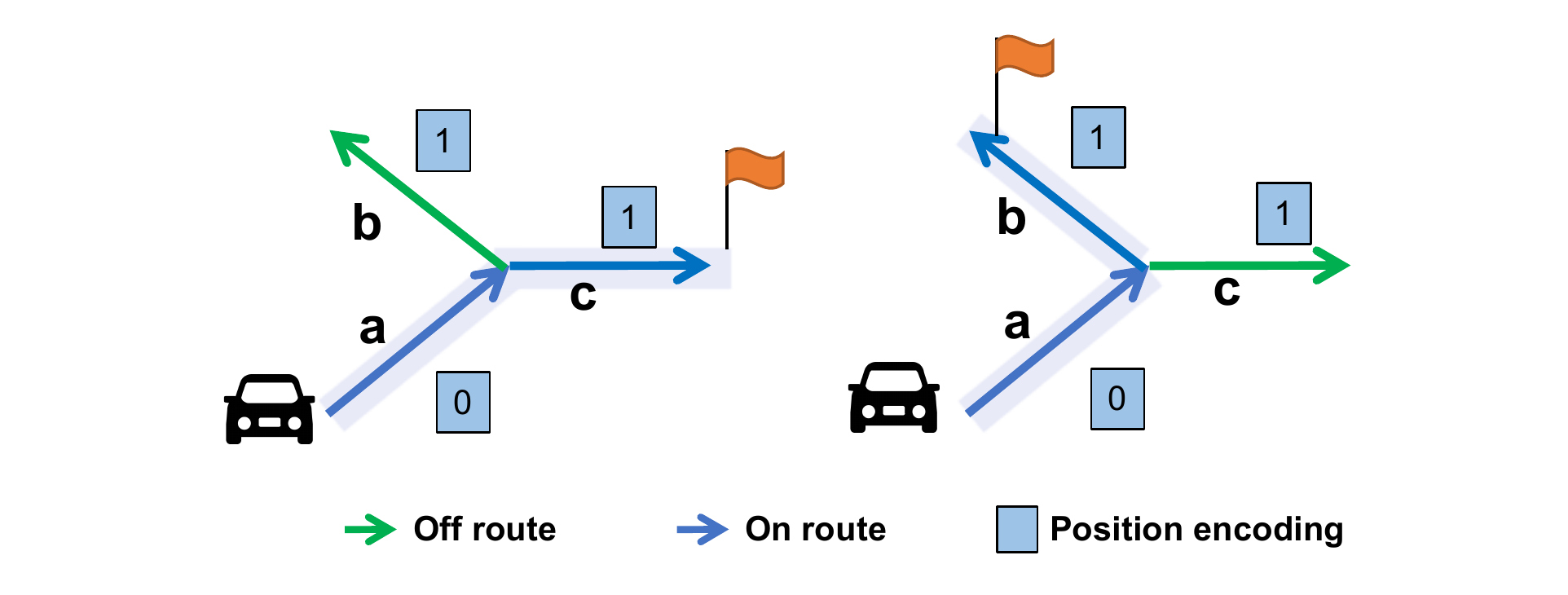}
		\caption{Graph transformer is unable to distinguish the two routes in case 1 and case 2, because it generates the same representations of link \textbf{a} in two cases.}
		\label{fig:route_identifier_1}
	\end{subfigure}
	\begin{subfigure}{1.0\columnwidth}
	    \setlength{\abovecaptionskip}{0.1cm}
        \setlength{\belowcaptionskip}{0.01cm}
		\centering
		\includegraphics[width=1.0\columnwidth,trim={0.0cm 0.5cm 0.0cm 0.0cm},clip]{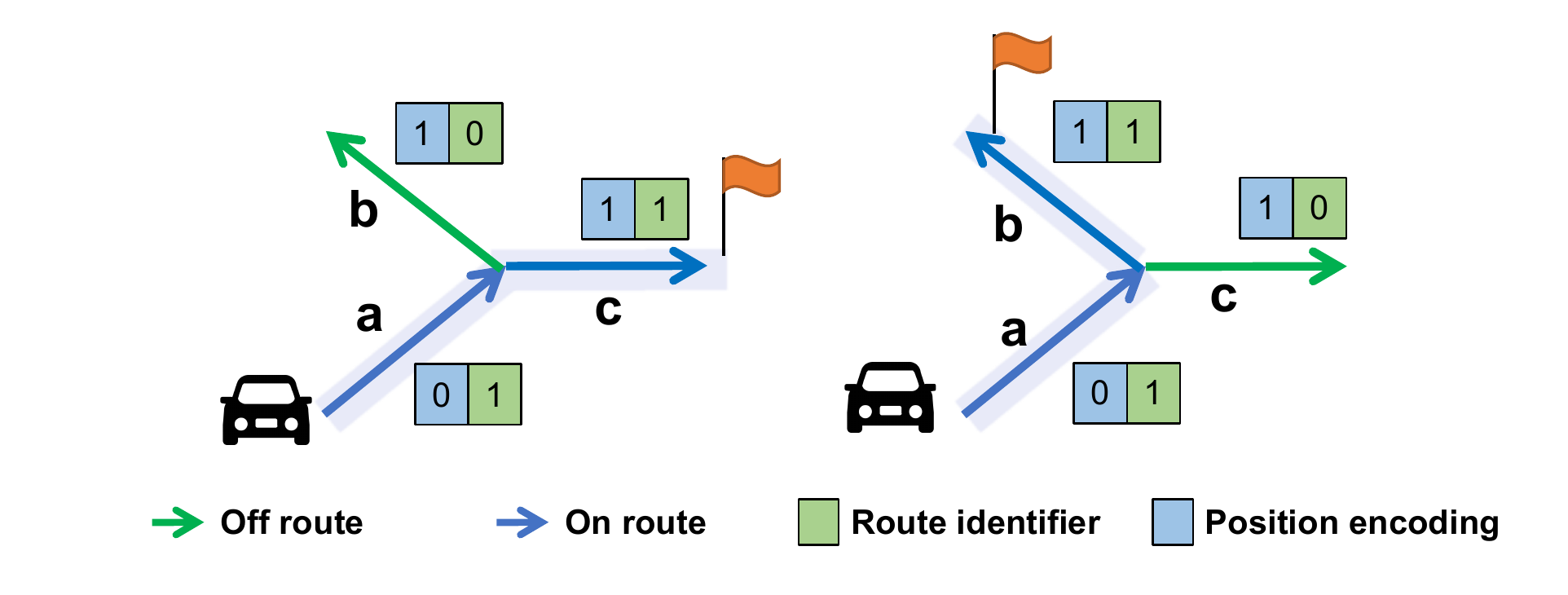}
		\caption{Route identifier is added to distinguish whether a link belongs to a route or not, which facilitates generating distinct representations of link \textbf{a} in two cases.}
		\label{fig:route_identifier_2}
	\end{subfigure} 
\caption{The necessity of introducing a route identifier.}
\vspace{-3mm}
\label{fig:route_emb_case}
\end{figure}

We believe that the route-level information is also crucial for ETA prediction, and we integrate it into link representations. However, it is more challenging to estimate the traffic condition of a link that is far from the origin of a route, because the traffic condition when a user arrives at this link is likely different from that when the user requests the ETA. In addition, the graph transformer is unable to identify whether a link belongs to a given route or not, making it difficult to generate distinct representations of a same link w.r.t. different routes, as showcased by Figure \ref{fig:route_identifier_1}.
To address these issues, we introduce two route-aware structural encoding methods, position encoding and route identifier, to improve the sensitivity on route information, as showcased by Figure \ref{fig:route_identifier_2}.

Specifically, the position encoding is designed to encode the order information of a link. Given a route, we calculate the shortest hop between the origin and a link in the road network as the link's position encoding. In this way, the position encoding of a link can be regarded as a gate to control the degree of dependency of the traffic condition when a user requests the ETA. 
In addition, a route identifier is added to distinguish whether a link belongs to a route or not, which enables the graph transformer to generate contextualized link representations.

\subsubsection{Integration}
To capture the local dependencies between links in the requested route, we use a 1-D convolution layer (Conv1D) with a window size of 3 \cite{krizhevsky2012imagenet} and take link representation vectors of the links in the route as input. Then, a multilayer perceptron (MLP) with ReLU \cite{agarap2018deep} as the activation function is employed, which takes the smoothed output from Conv1D as input and estimates the travel time of each link:
\begin{equation}
	[\hat{y}_1, \hat{y}_2, \cdots,\hat{y}_m]=\text{MLP}(\text{Conv1D}([\mathbf{h}_1, \mathbf{h}_2,\cdots, \mathbf{h}_m])),
\end{equation}
where $\hat{y}_i$ is the estimated travel time of the $i$-th link in the route.

The estimated travel times of all the links in the route are summed up as the estimated travel time of the entire route  $r=[l_1, l_2, \cdots, l_m]$:
\begin{equation}
	\hat{y}= \sum_i^{m} \hat{y}_i,
\end{equation}
where $\hat{y}$ denotes the estimated travel time of route $r$.

Multi-task learning is adopted to optimize the model parameters from both the link-level and the route-level. On the one hand, Huber loss \cite{huber1992robust} is used as the link-level loss function, which is defined as: 
\begin{equation}
	L_{link}(\hat{y}_i, y_i )= \left\{\begin{aligned}
		&\frac{1}{2}(\hat{y}_i - y_i)^2,& | \hat{y}_i - y_i | < \delta, \\
		&\delta( | \hat{y}_i - y_i |  - \frac{1}{\delta}),& \text{otherwise},\\
	\end{aligned} \right.
\end{equation}
where $\delta$ is a hyper-parameter to alleviate the impact of the outliers.
On the other hand, absolute percentage error (APE) is used as the loss function of the entire route $L_{route}$, which is defined as: 
\begin{equation}
	L_{route}(\hat{y}, y) = \frac{|\hat{y} - y|}{y}.
\end{equation}

We combine the link-level loss $L_{link}$ and the route-level loss $L_{route}$ to obtain the loss function $L$:
\begin{equation}
	L = \frac{1}{n}\sum_i^{n}(\frac{1}{m^{(i)}}\sum_{j=1}^{m^{(i)}}L_{link}(\hat{y}_j^{(i)}, y_j^{(i)}) + L_{route}(\hat{y}^{(i)}, y^{(i)})).
\end{equation}

We minimize $L$ to optimize the parameters of the model.

\section{Experiments}
\subsection{Experimental Setup}

\begin{table}[!ht]
   \setlength{\abovecaptionskip}{0.15cm}
   \setlength{\belowcaptionskip}{0.01cm}
    \setlength{\tabcolsep}{4.5pt}
	\caption{Statistics of the real-world datasets.}
	\label{tab:statistics}
	\begin{tabular}{c|cccc}
		\hline
		\multirow{2}{*}{\textbf{Dataset}} & \multirow{2}{*}{\textbf{\#Links}} & \textbf{\#Training} & \textbf{\#Test}   & \textbf{Average links} \\
		&                        & \textbf{routes}   & \textbf{routes} & \textbf{per route}     \\ \hline
		Beijing                  &       2,435,719                 &   20,067,736      &    4,153,587   &     96.87        \\
		Shanghai                 &      2,427,225                  &   23,844,299     &    5,280,030   &      87.98         \\
		Tianjin                  &     1,643,454           &     7,760,443        &    1,815,334    &      94.25        \\ \hline
	\end{tabular}
\end{table}

\begin{table*}[!t]
\setlength{\abovecaptionskip}{0.1cm}
\setlength{\belowcaptionskip}{0.01cm}
	\caption{Performance of DuETA and the baseline methods for ETA prediction on three real-world datasets.}
	\label{tab:overall}
	\begin{tabular}{@{}c|ccc|ccc|ccc@{}}
		\toprule
		& \multicolumn{3}{c|}{\textbf{Beijing}} & \multicolumn{3}{c|}{\textbf{Shanghai}} & \multicolumn{3}{c}{\textbf{Tianjin}} \\
		\textbf{Method} & \textbf{MAE (sec)}      & \textbf{RMSE (sec)}  & \textbf{MAPE (\%)}   & \textbf{MAE (sec)}      & \textbf{RMSE (sec)}   & \textbf{MAPE (\%)}   & \textbf{MAE (sec)}     & \textbf{RMSE (sec)} & \textbf{MAPE (\%)}    \\ \midrule
		AVG   & 367.37            & 750.68      & 41.25      & 312.22            & 575.82       & 39.74      & 316.47           & 567.34   & 35.94         \\ \hline
		STANN           &     206.18      &    450.57 &  24.37          &   184.46  &    369.88     &    24.32     &      189.01            &        395.59          &    21.48               \\
		DCRNN           &      204.65   &   444.10       &   24.81   &       183.52     &      367.08      &  24.27     &            188.28       &         395.15         &          21.40         \\ \hline
		DeepTravel      & 183.87           & 410.92    & 21.67       & 170.09      & 328.35 & 25.10             & 178.66           & 386.33   & 19.62        \\
		ConSTGAT        & 181.91           & 401.21 & 22.03           & 158.33            & 319.62   & 21.05        & 176.61      &  381.49 & 19.59       \\ \midrule
		DuETA           & \textbf{178.39}            & \textbf{399.12}   & \textbf{21.22}         & \textbf{155.85}            & \textbf{315.82}       & \textbf{20.83}       & \textbf{171.42}           & \textbf{370.59} & \textbf{19.50}          \\ \bottomrule
	\end{tabular}
\vspace{-3mm}
\end{table*}

We evaluate DuETA against several competitive baseline methods on three real-world, large-scale datasets, including Beijing, Shanghai, and Tianjin. These metropolises are the largest cities in China, which contain millions of links individually. The datasets are sampled from Baidu Maps, which consist of tens of millions of routes and their corresponding travel times ranging from Oct. 10th to Nov. 20th, 2021. 
The selected period minimizes the impact of the COVID-19 pandemic on traffic \cite{huang2020understanding}.
The data of the first four weeks are used for training, while the data of the last week are used for evaluation. The statistics of the datasets are shown in Table \ref{tab:statistics}. 

Three widely-used metrics for evaluating ETA prediction, including mean average error (MAE), root mean square error (RMSE), and mean absolute percentage error (MAPE), are used to evaluate the methods, which are defined as:

\begin{equation}
MAE(\hat{y}^{(i)}, y^{(i)}) =\frac{1}{n} \sum_{i=1}^{n}|\hat{y}^{(i)} -  y^{(i)} |,
\end{equation}

\begin{equation}
RMSE(\hat{y}^{(i)}, y^{(i)})=\sqrt{\frac{1}{n}\sum_{i=1}^{n}(\hat{y}^{(i)} -  y^{(i)} )^ 2},
\end{equation}

\begin{equation}
MAPE(\hat{y}^{(i)}, y^{(i)}) =\frac{1}{n} \sum_{i=1}^{n}\frac{|\hat{y}^{(i)} -  y^{(i)} |}{ y^{(i)}}.
\end{equation}

\subsection{Baselines}
We compare DuETA against the following five baseline methods. 
\begin{itemize}
	\item \textbf{AVG}. For each dataset, we calculate the average travel times of each link at each time slot according to the records in the training set. Given a request $req=(r,s)$, we sum up the statistical travel times of all the links in route $r$ with the departure time $s$ to estimate the travel time of $r$.
	\item \textbf{STANN} \cite{he2018stann}. STANN is an STGNN, which encodes the spatial and temporal traffic information by attention mechanism and LSTM. The standard STANN assumes that all links in the road network are connected, making it computationally impractical for large-scale datasets. Therefore, we revise it to consider only the spatially connected links.
	\item \textbf{DCRNN} \cite{li2018diffusion}. DCRNN is also an STGNN, which first captures the spatial information by graph convolution network and then captures temporal information by LSTM.
	\item \textbf{DeepTravel}  \cite{zhang2018deeptravel}. DeepTravel is an end-to-end method, where the spatial and temporal features are extracted and taken as the input of a bidirectional LSTM to estimate the travel time. We re-implement DeepTravel based on the road network instead of the original spatial grids.
	\item \textbf{ConSTGAT} \cite{fang2020constgat}.  ConSTGAT is our previously deployed end-to-end ETA prediction model at Baidu Maps, which models the joint relations of spatial and temporal information as well as the contextual information of a route.
\end{itemize}

All methods are implemented using PaddlePaddle, an open-source deep-learning platform maintained by Baidu. We set the embedding size and the hidden size of DuETA to be 32. The number of attention heads $C$ is set to be 8. The same settings are applied to the baseline methods, for a fair comparison. The model parameters are optimized by Adam optimizer \cite{kingma2014adam} with a learning rate of $3\times10^{-5}$. The model hyperparameters are tuned according to the validation performance, online serving latency, and model capacity. 

\begin{figure}[!bp]
\vspace{-5mm}
\setlength{\abovecaptionskip}{0.15cm}
   \centering
    \includegraphics[width=0.95\columnwidth,trim={0.75cm 0.7cm 0.75cm 0.7cm},clip]{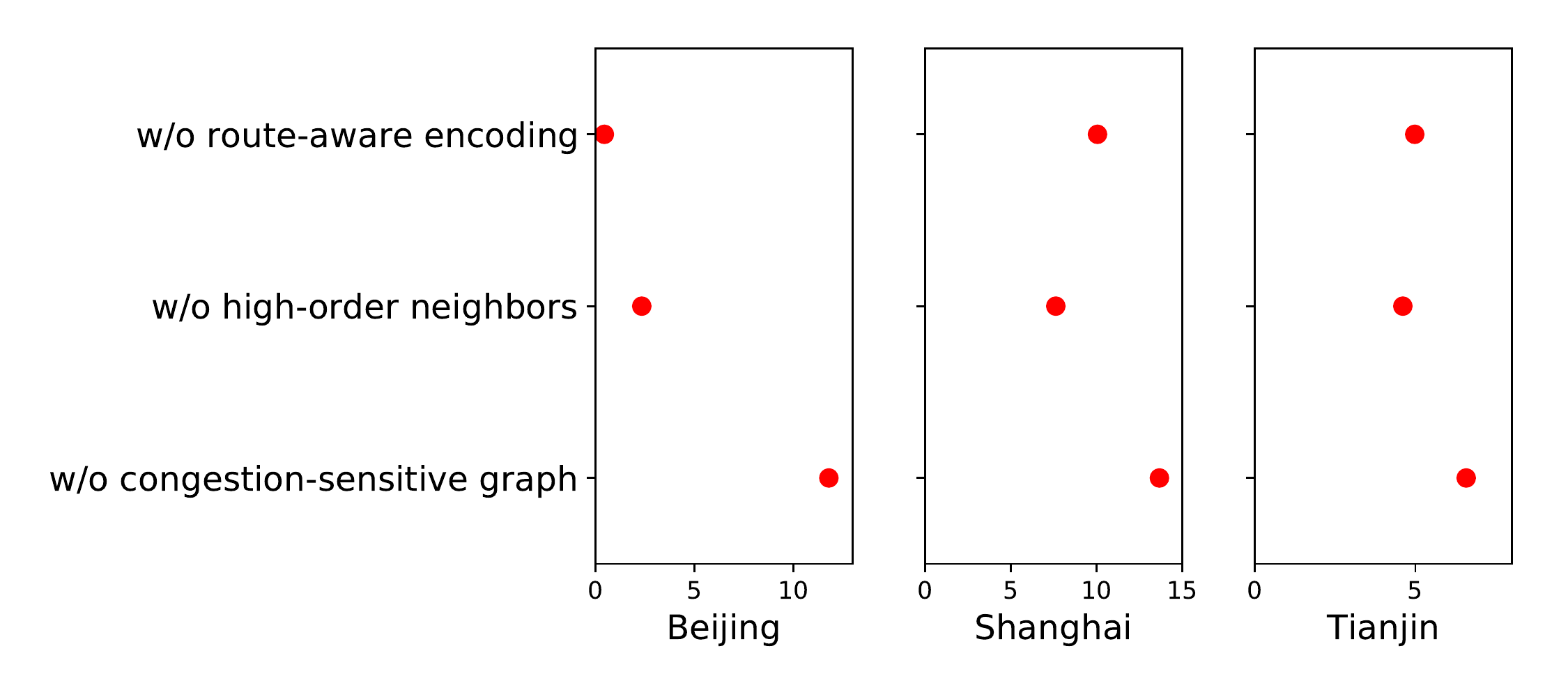}
   \caption{The $-\Delta$RMSE of ablation versions of DuETA.}
   \label{fig:ablation}
\end{figure}

\begin{figure*}[!t]
\captionsetup[subfigure]{labelformat=empty}
\setlength{\abovecaptionskip}{0.12cm}
	\centering
	\begin{subfigure}[t]{0.33\linewidth}
	    \setlength{\abovecaptionskip}{0.1cm}
        \setlength{\belowcaptionskip}{0.01cm}
		\centering
		\includegraphics[width=1.0\columnwidth,trim={0.0cm 0.0cm 0.0cm 0.7cm},clip]{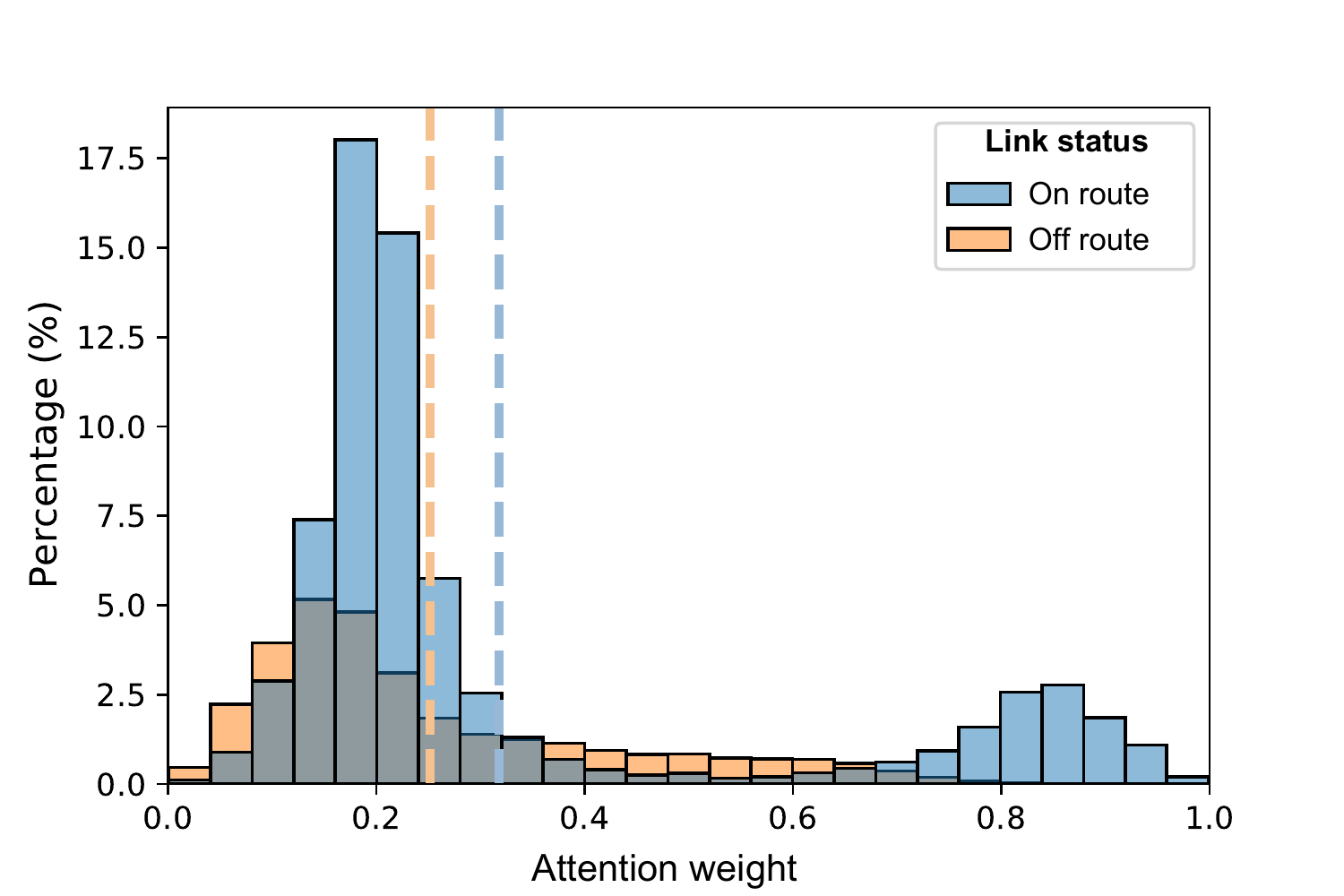}
		\caption{(a) Complete DuETA in Beijing.}
		\label{fig:beijing_v16}
	\end{subfigure}
    \begin{subfigure}[t]{0.33\linewidth}
	    \setlength{\abovecaptionskip}{0.1cm}
        \setlength{\belowcaptionskip}{0.01cm}
		\centering
		\includegraphics[width=1.0\columnwidth,trim={0.0cm 0.0cm 0.0cm 0.7cm},clip]{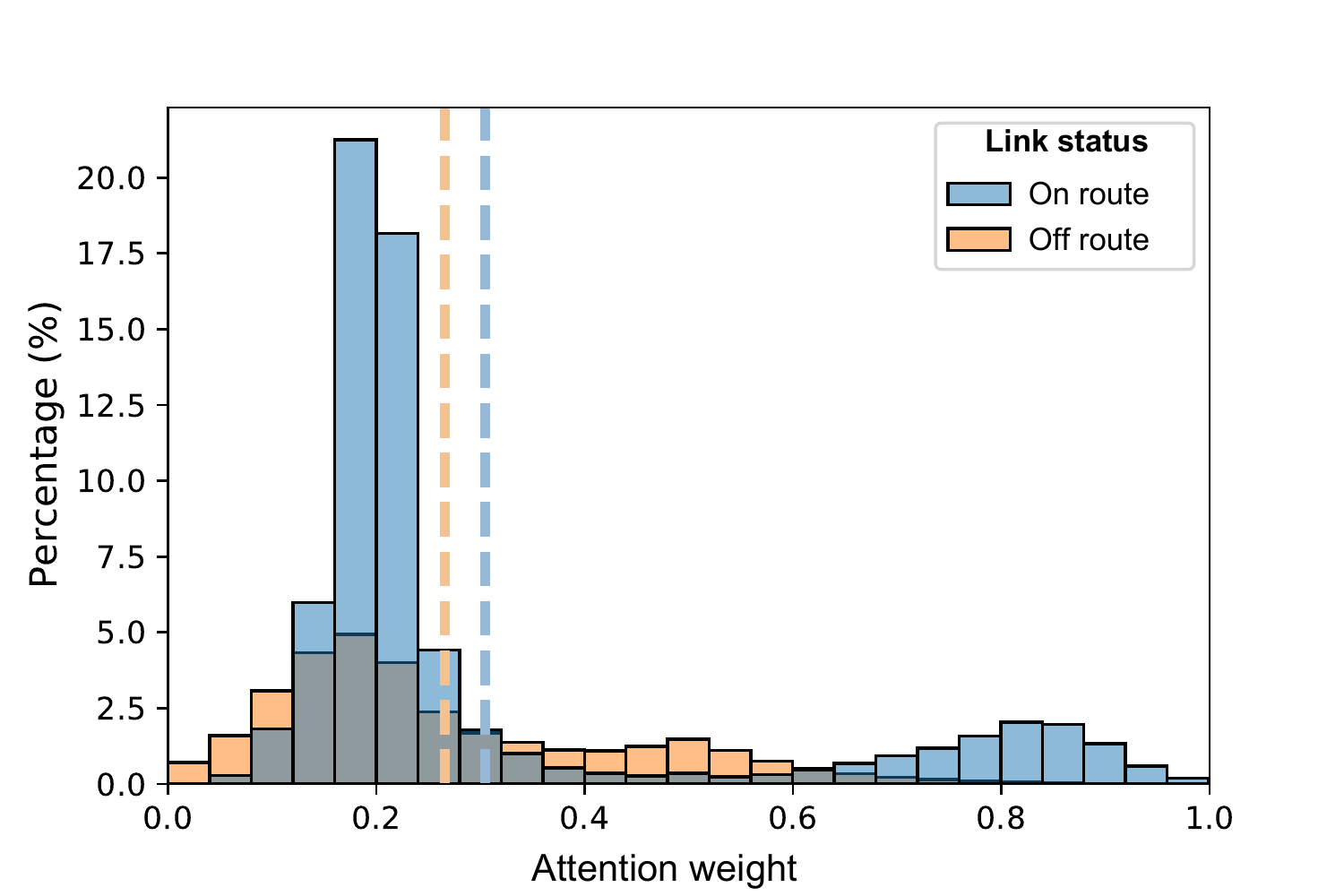}
		\caption{(c) Complete DuETA in Shanghai.}
		\label{fig:shanghai_v16}
	\end{subfigure} 
	\begin{subfigure}[t]{0.33\linewidth}
	    \setlength{\abovecaptionskip}{0.1cm}
        \setlength{\belowcaptionskip}{0.01cm}
		\centering
		\includegraphics[width=1.0\columnwidth,trim={0.0cm 0.0cm 0.0cm 0.7cm},clip]{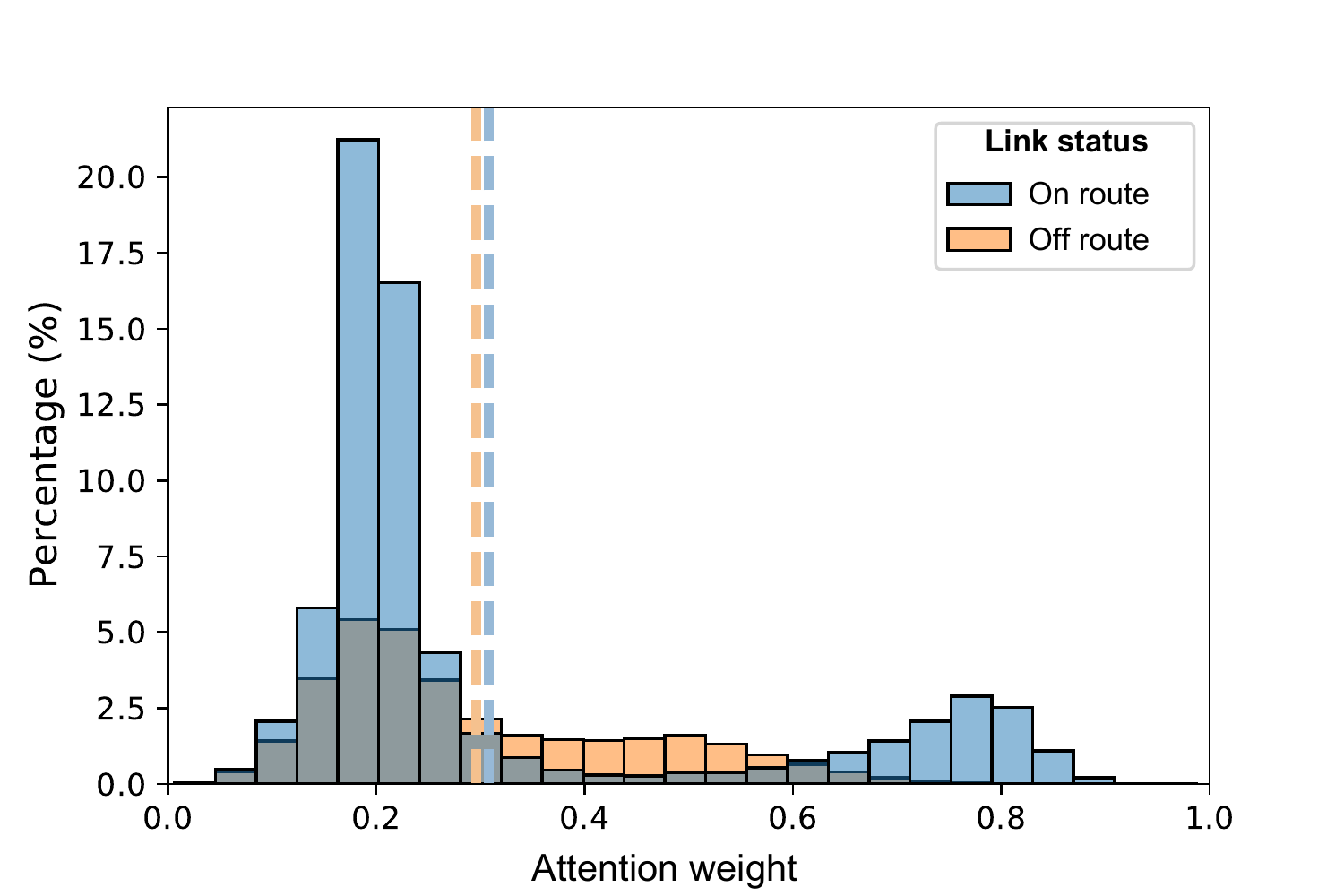}
		\caption{(e) Complete DuETA in Tianjin.}
		\label{fig:tianjin_v16}
	\end{subfigure} 
	\begin{subfigure}[t]{0.33\linewidth}
        \setlength{\abovecaptionskip}{0.1cm}
        \setlength{\belowcaptionskip}{0.01cm}
    	\centering
    	\includegraphics[width=1.0\columnwidth,trim={0.0cm 0.0cm 0.0cm 0.7cm},clip]{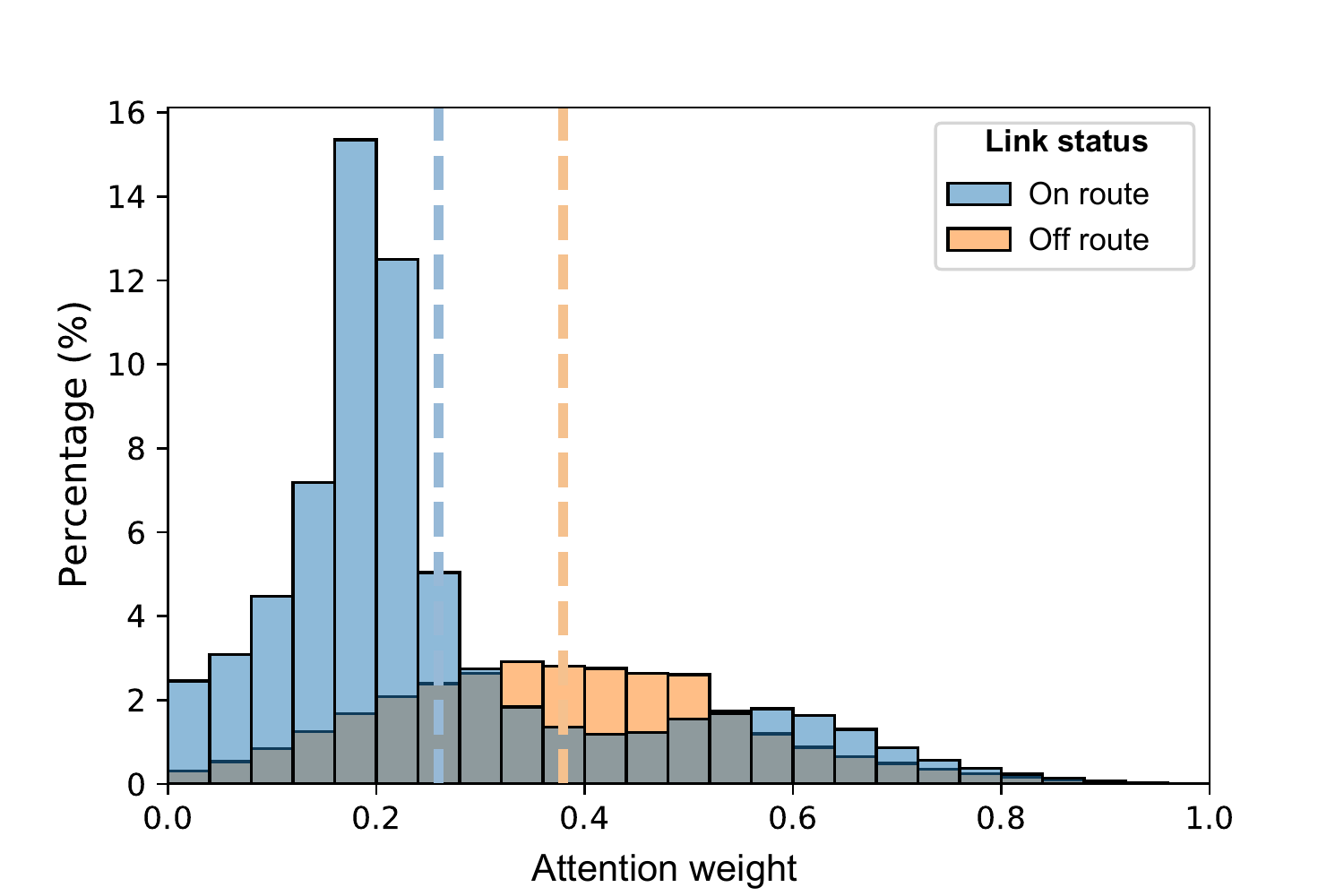}
    	\caption{(b) DuETA w/o route identifier in Beijing.}
    	\label{fig:beijing_v10}
	\end{subfigure}
    \begin{subfigure}[t]{0.33\linewidth}
        \setlength{\abovecaptionskip}{0.1cm}
        \setlength{\belowcaptionskip}{0.01cm}
    	\centering
    	\includegraphics[width=1.0\columnwidth,trim={0.0cm 0.0cm 0.0cm 0.7cm},clip]{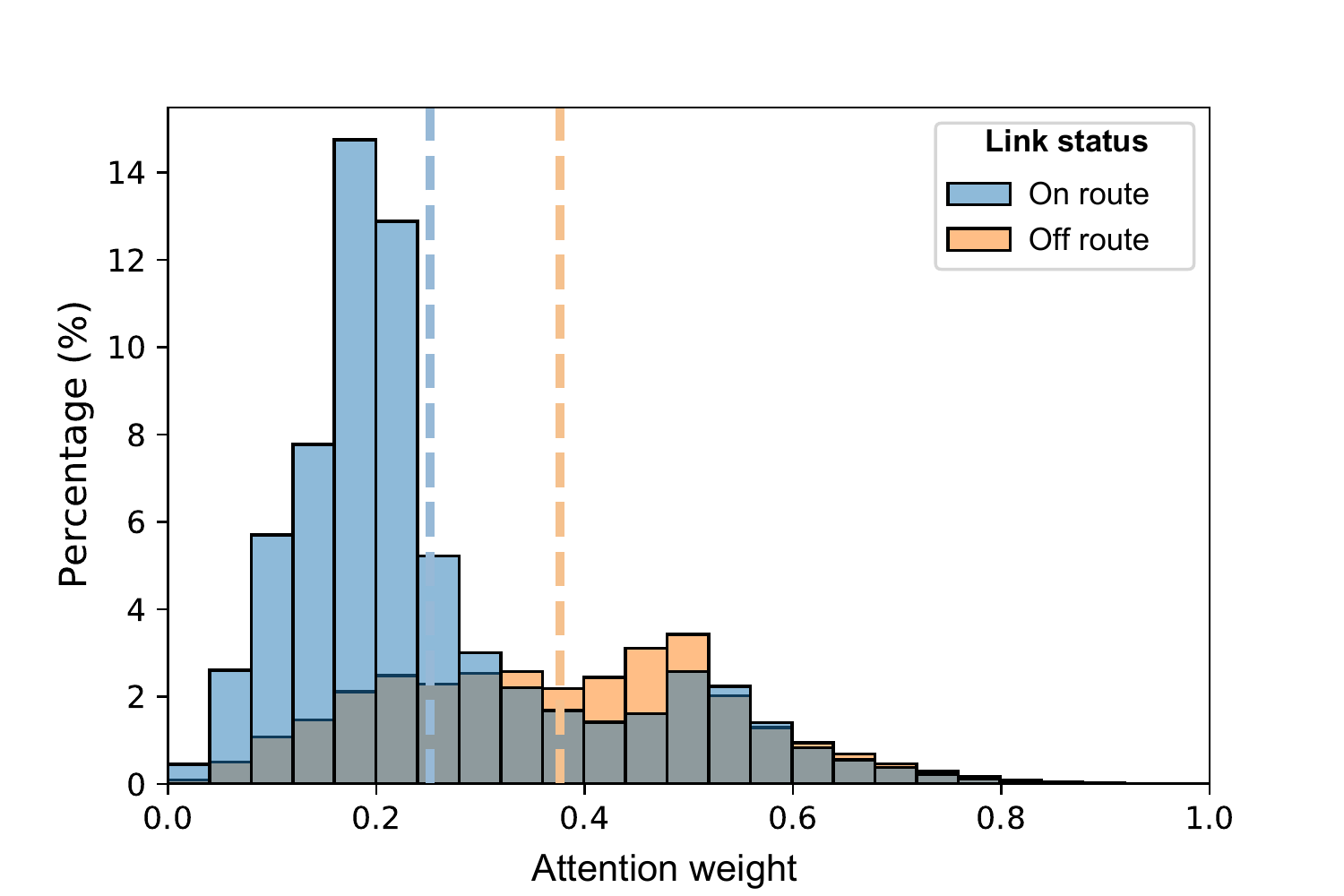}
    	\caption{(d) DuETA w/o route identifier in Shanghai.}
    	\label{fig:shanghai_10}
    \end{subfigure} 
    \begin{subfigure}[t]{0.33\linewidth}
	    \setlength{\abovecaptionskip}{0.1cm}
        \setlength{\belowcaptionskip}{0.01cm}
		\centering
		\includegraphics[width=1.0\columnwidth,trim={0.0cm 0.0cm 0.0cm 0.7cm},clip]{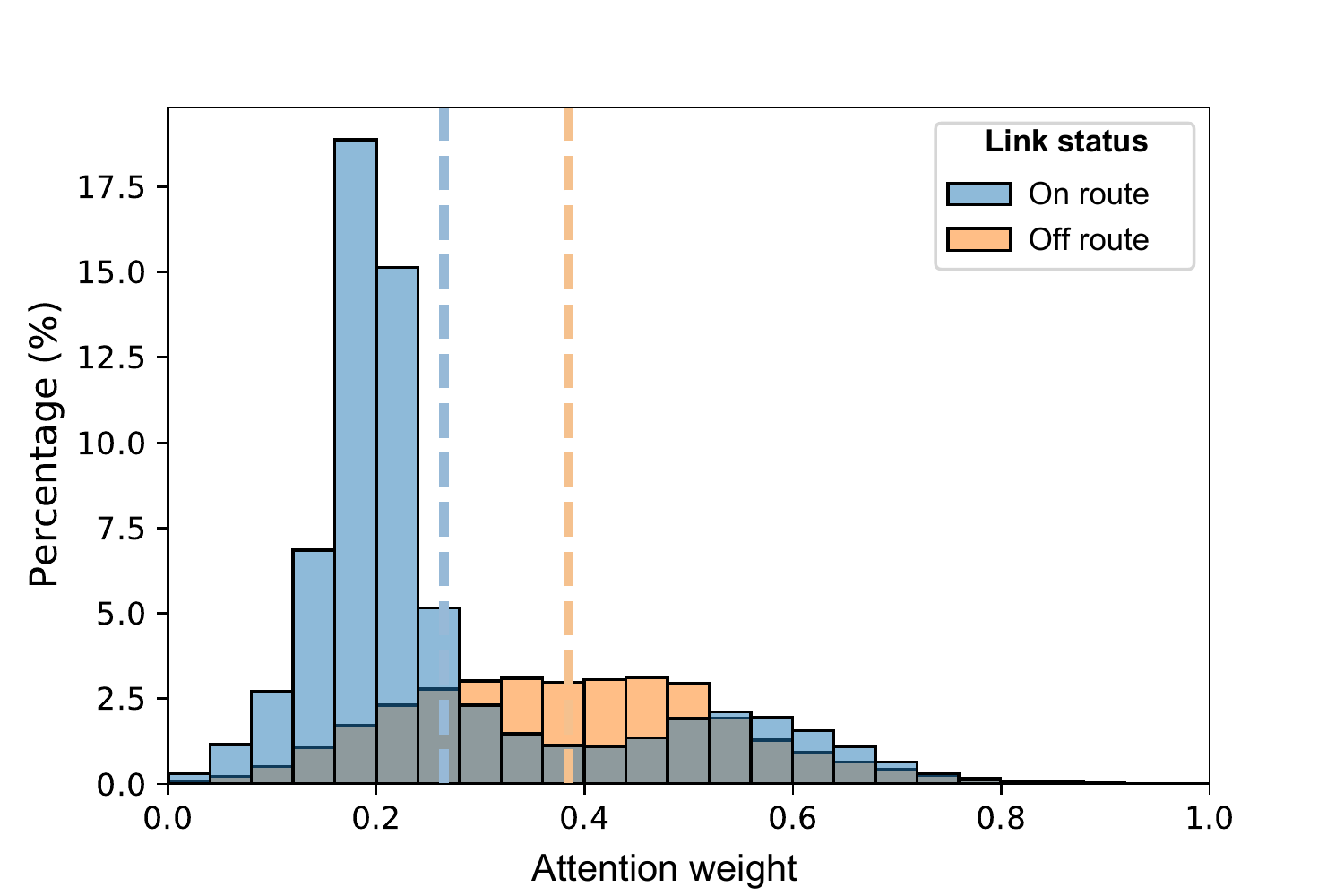}
		\caption{(f) DuETA w/o route identifier in Tianjin.}
		\label{fig:tianjin_v10}
	\end{subfigure} 
\caption{Visualization of the attention weight distributions of the complete DuETA and the ablative DuETA without a route identifier. The vertical dotted line denotes the mean value of the attention weights.}
\label{fig:attn_visualize}
\vspace{-2mm}
\end{figure*}

\subsection{Results and Analysis}
\subsubsection{Overall Performance}
We conducted offline tests to compare DuETA with multiple strong baselines to demonstrate the superiority of DuETA. Table \ref{tab:overall} shows the experimental results. Boldface indicates the best score w.r.t. each metric. From the results, we have the following observations.

(1) The performance of the naive baseline AVG is the worst, as it makes a simple assumption that the traffic condition of each link at each time slot is constant.

(2) STANN and DCRNN significantly outperform AVG by a large margin in terms of all metrics on three datasets. The main reason is that they both take into account the spatial and temporal information of the traffic conditions.

(3) DeepTravel and ConSTGAT significantly improve over both STANN and DCRNN on three datasets. The improvements are mainly due to two reasons: End-to-end methods are more effective than the segment-based methods, and the correlations of spatial and temporal information are jointly modeled.

(4) Results show that DuETA significantly outperforms all baselines, which demonstrates that modeling traffic congestion propagation patterns is able to significantly benefit the performance of ETA prediction.
The main reason is two-fold. On the one hand, DuETA is more sensitive to long-distance traffic congestion, because the introduction of high-order neighbors enables DuETA to bridge the gap between links that are not directly connected yet are highly correlated.
On the other hand, the cumulative effect of delay variations over time caused by traffic events on the road network can be alleviated by the high efficiency of traffic congestion pattern modeling.

\subsubsection{Ablation Studies}
We perform ablation experiments to understand the relative importance of different components of DuETA.

First, we examine the impact of different components of DuETA, including the route-aware graph transformer and the congestion-sensitive graph.
The ablation results in Figure \ref{fig:ablation} show that removing both components hurts performance significantly in all three cities.
This demonstrates the significance of both components in improving ETA prediction performance.
Moreover, removing the high-order neighbors (DuETA w/o high-order neighbors) leads to significant drops in performance on all datasets. This confirms the effectiveness of the long-distance associations between links for ETA prediction.

Second, we study the effect of the route-aware graph transformer.
To learn the long-distance associations between links, DuETA adopts a route-aware graph transformer that encodes route identifier and position information. 
To obtain an understanding of the effect of the route identifier, we visualize the distributions of the attention weights in Figure \ref{fig:attn_visualize}. The sub-figures at the top of Figure \ref{fig:attn_visualize} show the distributions of the attention weights of the complete DuETA, while the sub-figures at the bottom show those of the ablative DuETA without a route identifier (abbr. ``DuETA w/o RI''). From the results, we have the following observations.
For the links that are in the travel routes (abbr. ``On route'' in Figure \ref{fig:attn_visualize}), the averaged attention weights of the complete DuETA are larger than those of ``DuETA w/o RI'' in all three cities. 
By contrast, for the links that are not in the travel routes (abbr. ``Off route'' in Figure \ref{fig:attn_visualize}), the averaged attention weights of the complete DuETA are smaller than those of ``DuETA w/o RI''. 
This demonstrates that the introduction of a route identifier enables DuETA to pay more attention to the links that are in the travel routes.

\begin{figure}[!htbp]
\setlength{\abovecaptionskip}{0.1cm}
   \centering
    \includegraphics[width=1.0\columnwidth,trim={0.45cm 0.6cm 0.45cm 0.3cm},clip]{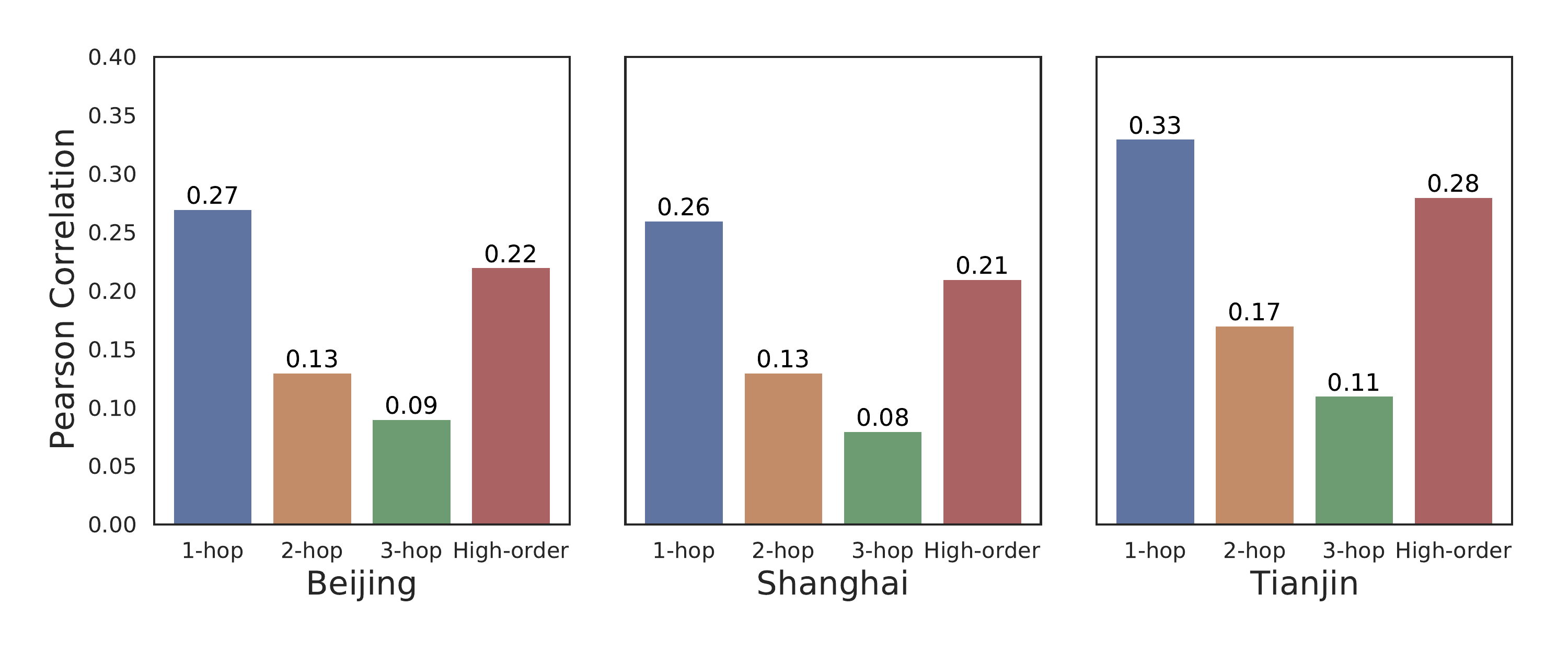}
   \caption{Averaged Pearson correlation coefficients of traffic patterns between the links and their neighbors.}
   \label{fig:road_network_analysis}
\vspace{-4mm}
\end{figure}

Third, we study the effect of a congestion-sensitive graph. To enable DuETA to capture the interactions between links that are spatially distant but highly correlated with traffic conditions, we construct a congestion-sensitive graph based on the correlations of traffic patterns. Figure \ref{fig:road_network_analysis} presents the Pearson correlation coefficients of the traffic patterns between the links and their neighbors. 
It can be seen that the correlations between the links and their first-order neighbors are the strongest among all kinds of neighbors.
This is intuitive, because the traffic flows on spatially adjacent and closely connected links can easily influence each other.
Although the correlations between the links and their second-order and third-order neighbors are relatively lower than that of first-order neighbors, they can also contribute to improving the performance of ETA prediction.
Moreover, the average Pearson correlation coefficients of our selected high-order neighbors is much higher than those of the second-order and third-order neighbors, which demonstrates the significant role of high-order neighbors in improving ETA prediction performance. 
To further investigate the impact of high-order neighbors for traffic congestion pattern modeling, we examine the relative improvements of high-order neighbors in cases of traffic congestion\footnote{Here, traffic congestion is defined as speed $\leq$ 10 km/h.} and normal traffic. Table \ref{tab:traffic_congestion} shows the relative improvements of high-order neighbors in cases of different traffic conditions. Results show that the improvements achieved by DuETA in case of traffic congestion are much higher than those of normal traffic. This further verifies the effectiveness of high-order neighbors in improving ETA prediction.

\begin{table}[!th]
\vspace{-1mm}
\setlength{\abovecaptionskip}{0.15cm}
\caption{The relative improvements (MAPE) of high-order neighbors in different traffic conditions. }
\label{tab:traffic_congestion}
\centering
\begin{tabular}{c|c|c}
\hline
             &  \textbf{Congestion} & \textbf{Normal} \\ \hline
\textbf{Beijing}  & 12.2\%             & 6.6\%                      \\ \hline
\textbf{Shanghai} & 12.0\%             & 6.8\%                      \\ \hline
\textbf{Tianjin}  & 17.0\%              & 5.9\%                      \\ \hline
\end{tabular}
\vspace{-2mm}
\end{table}

\subsection{Practical Applicability}
\begin{figure}[!ht]
\setlength{\abovecaptionskip}{0.1cm}
\setlength{\belowcaptionskip}{0.01cm}
	\centering
	\begin{subfigure}{1.0\columnwidth}
	    \setlength{\abovecaptionskip}{0.06cm}
        \setlength{\belowcaptionskip}{0.1cm}
		\centering
		\includegraphics[width=1.0\columnwidth,trim={0.7cm 1.7cm 0.7cm 0.7cm},clip]{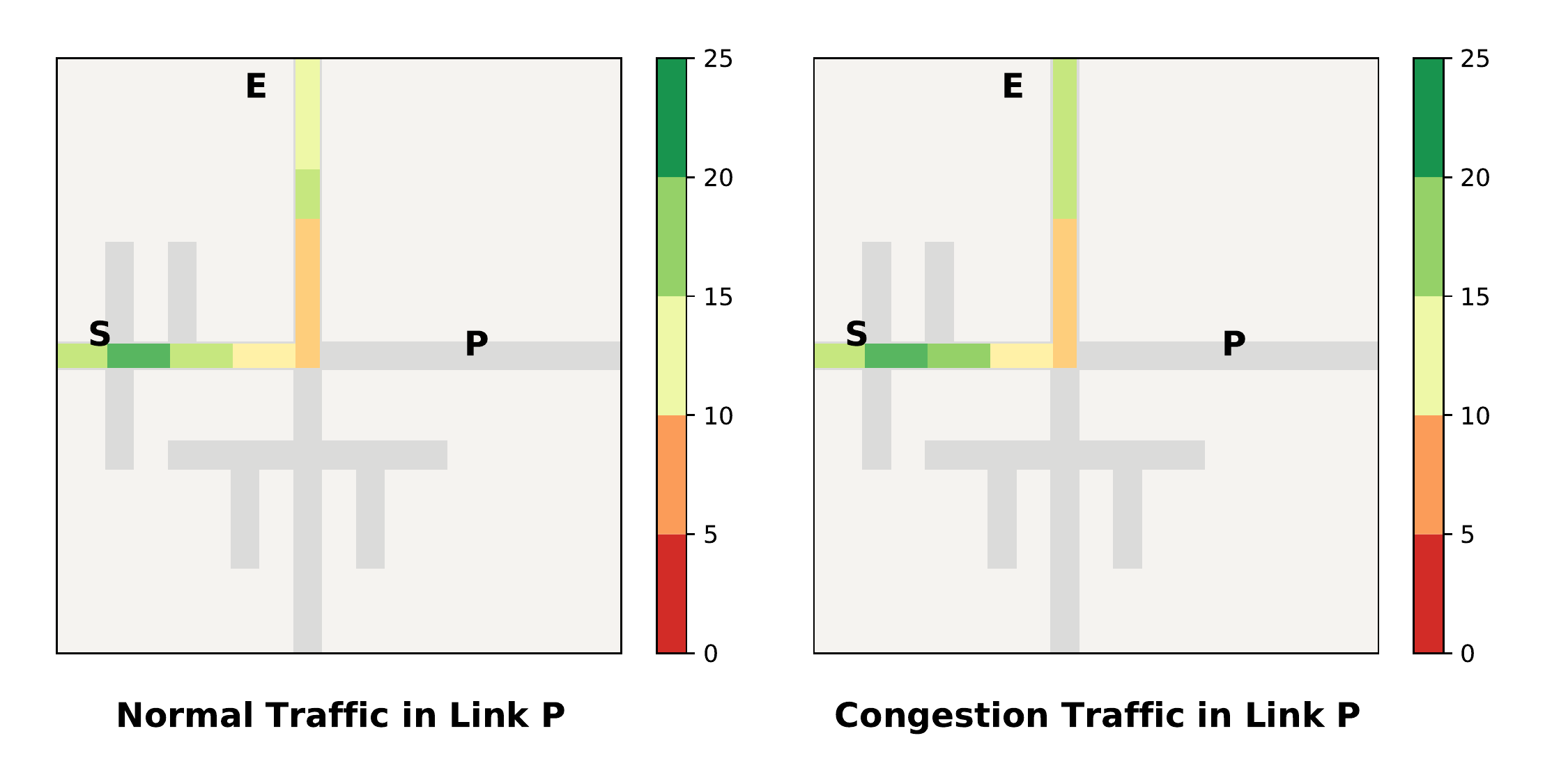}
		\includegraphics[width=1.0\columnwidth,trim={0.7cm 0.3cm 0.7cm 0.4cm},clip]{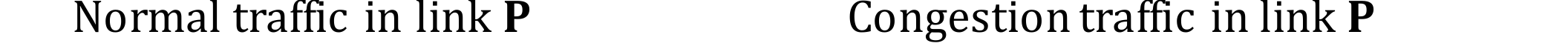}
		\caption{Future traffic conditions predicted by ConSTGAT.}
		\label{fig:pred_case1}
	\end{subfigure}
	
	\begin{subfigure}{1.0\columnwidth}
	    \setlength{\abovecaptionskip}{0.06cm}
        \setlength{\belowcaptionskip}{0.01cm}
		\centering
		\includegraphics[width=1.0\columnwidth,trim={0.7cm 1.7cm 0.7cm 0.7cm},clip]{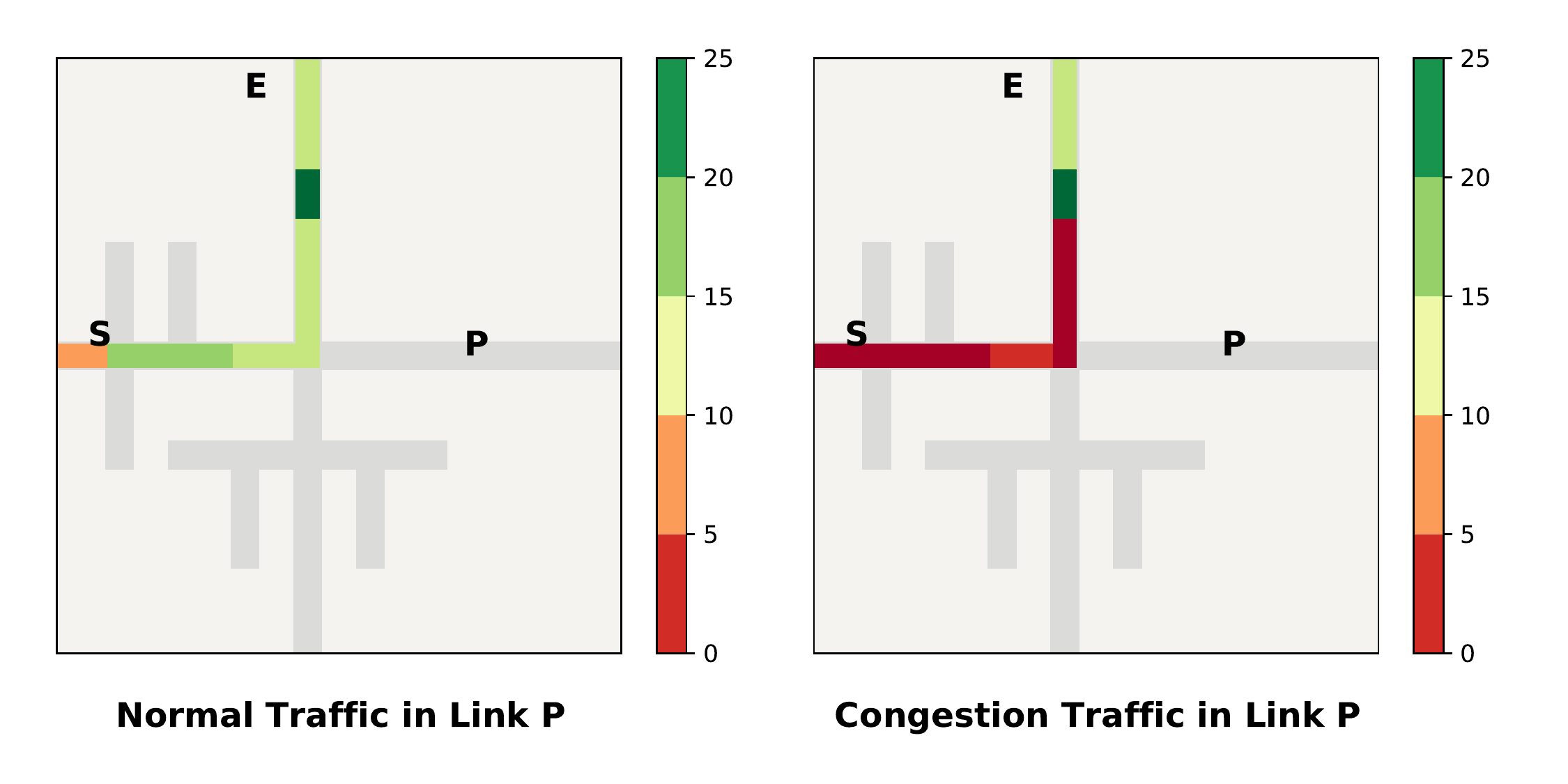}
		\includegraphics[width=1.0\columnwidth,trim={0.7cm 0.3cm 0.7cm 0.4cm},clip]{figures/f9-note}
		\caption{Future traffic conditions predicted by DuETA.}
		\label{fig:pred_case2}
	\end{subfigure} 
\caption{Comparison of DuETA and ConSTGAT in modeling traffic congestion propagation. The estimated travel speed (km/h) of each link is marked with a distinctive color.}
\label{fig:pred_cases}
\vspace{-4mm}
\end{figure}

Before deploying DuETA in production, we conducted a case study to validate the practicability of DuETA, as well as to verify the superiority of DuETA in traffic congestion propagation modeling over the previously deployed model, ConSTGAT.
Specifically, given a requested travel route $r$ that consists of the origin link \textbf{S}, the destination link \textbf{E}, and the links in between \textbf{S} and \textbf{E}, we compare the future traffic conditions of the links in $r$ predicted by ConSTGAT and DuETA when congestion occurs in a distant link \textbf{P}.
Figure \ref{fig:pred_cases} shows two cases of future traffic conditions predicted by ConSTGAT and DuETA.
From the results, we make the following observations.
First, the sub-figures at the left present the results predicted by ConSTGAT and DuETA when the traffic condition of link \textbf{P} is normal. 
We can see that the future traffic conditions of the links in $r$ predicted by both models are mostly normal without any congestion.
Second, the sub-figures at the right illustrate the results predicted by ConSTGAT and DuETA when severe congestion occurs in link \textbf{P}. 
The results in Figure \ref{fig:pred_case2} clearly show that the traffic conditions of most links (predicted by DuETA) in $r$ are propagated from the congestion in distant link \textbf{P}, while the traffic conditions of these links (predicted by ConSTGAT) are still normal, as shown in Figure \ref{fig:pred_case1}.
This demonstrates that DuETA is more sensitive to distant congestion than ConSTGAT, and it confirms the superiority of DuETA for the problem of traffic congestion propagation modeling.

\subsection{Online Evaluation}
\begin{figure*}[!ht]
\setlength{\abovecaptionskip}{0.1cm}
\setlength{\belowcaptionskip}{0.01cm}
	\centering
	\begin{subfigure}{0.33\linewidth}
	    \setlength{\abovecaptionskip}{0.1cm}
        \setlength{\belowcaptionskip}{0.01cm}
		\centering
		\includegraphics[width=1.0\columnwidth,trim={0.0cm 0.0cm 0.0cm 0.0cm},clip]{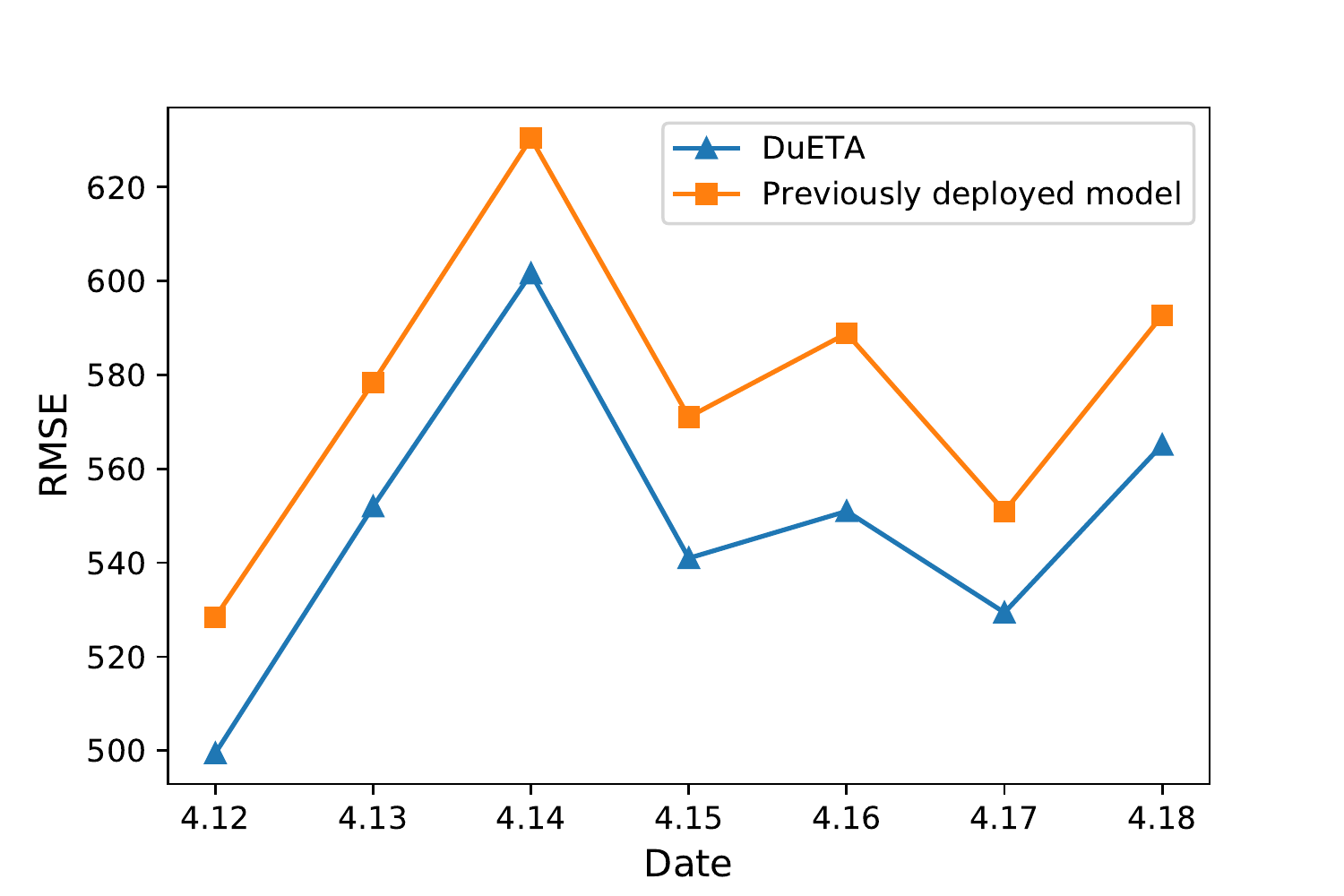}
		\caption{Overall performance.}
		\label{fig:online_case1}
	\end{subfigure}
	\begin{subfigure}{0.33\linewidth}
	    \setlength{\abovecaptionskip}{0.1cm}
        \setlength{\belowcaptionskip}{0.01cm}
		\centering
		\includegraphics[width=1.0\columnwidth,trim={0.0cm 0.0cm 0.0cm 0.0cm},clip]{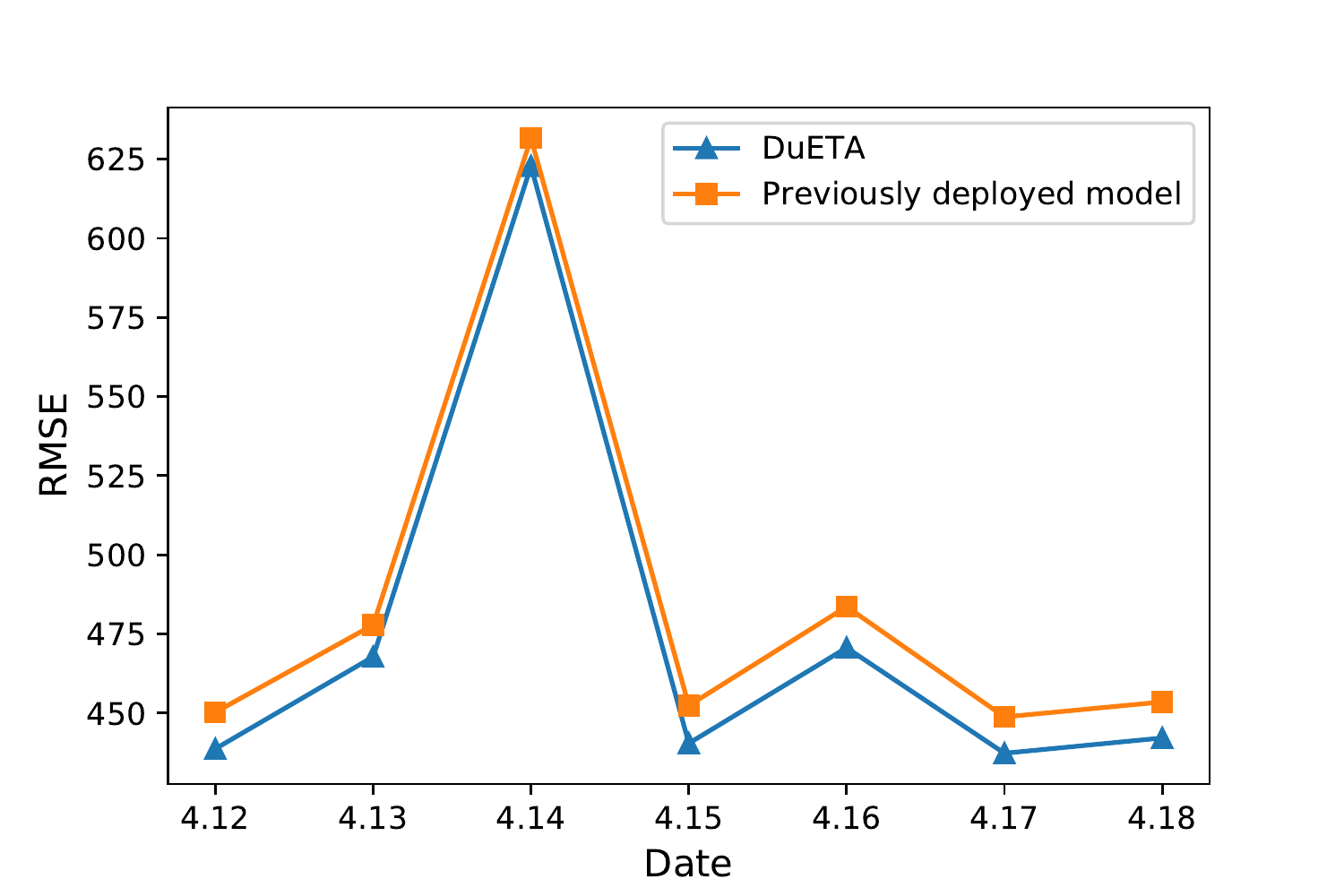}
		\caption{Short travel routes ($<$3km).}
		\label{fig:online_case2}
	\end{subfigure}
	\begin{subfigure}{0.33\linewidth}
	    \setlength{\abovecaptionskip}{0.1cm}
        \setlength{\belowcaptionskip}{0.01cm}
		\centering
		\includegraphics[width=1.0\columnwidth,trim={0.0cm 0.0cm 0.0cm 0.0cm},clip]{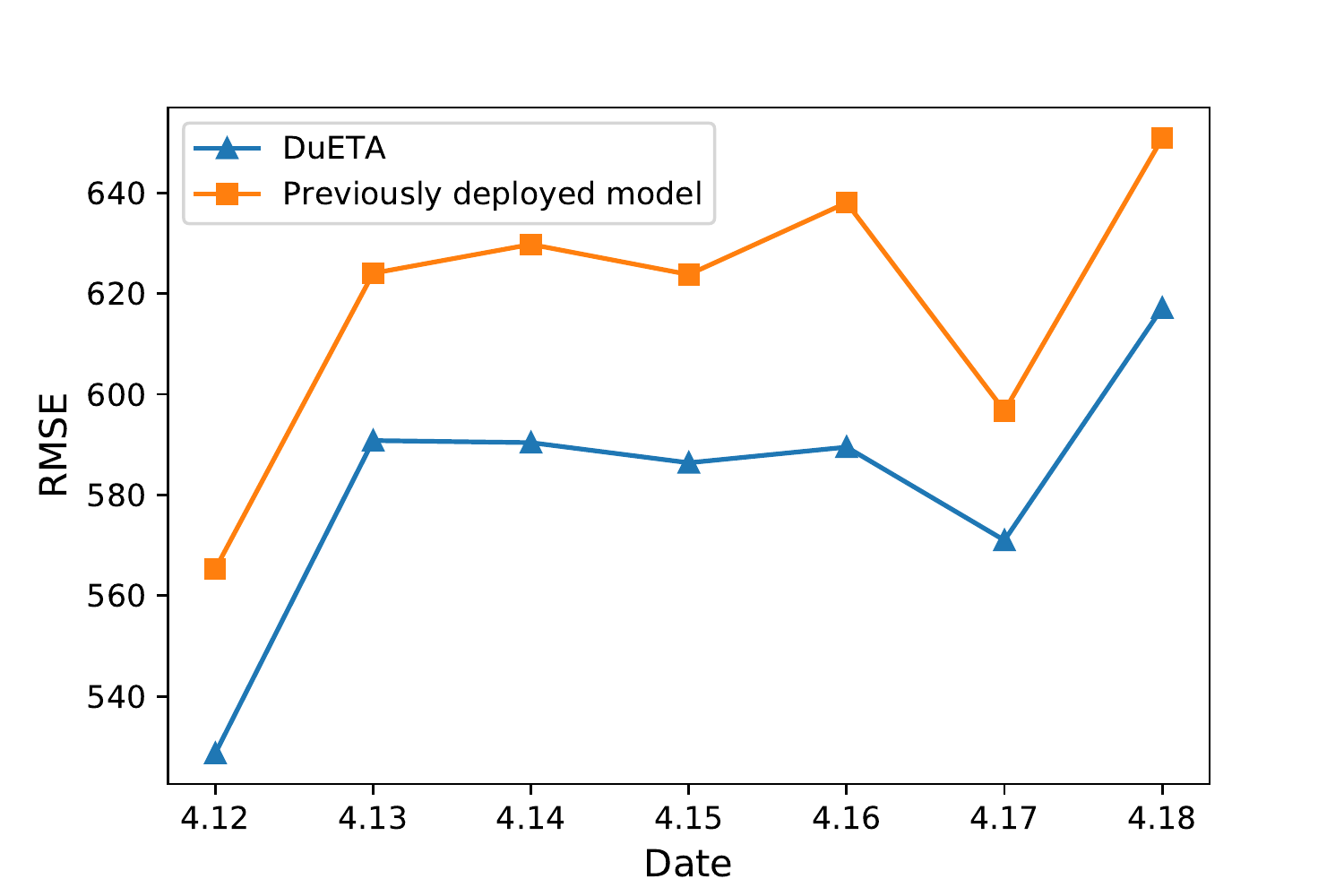}
		\caption{Long travel routes ($\ge$3km).}
		\label{fig:online_case3}
	\end{subfigure}
	\begin{subfigure}{0.33\linewidth}
	    \setlength{\abovecaptionskip}{0.1cm}
        \setlength{\belowcaptionskip}{0.01cm}
		\centering
		\includegraphics[width=1.0\columnwidth,trim={0.0cm 0.0cm 0.0cm 0.0cm},clip]{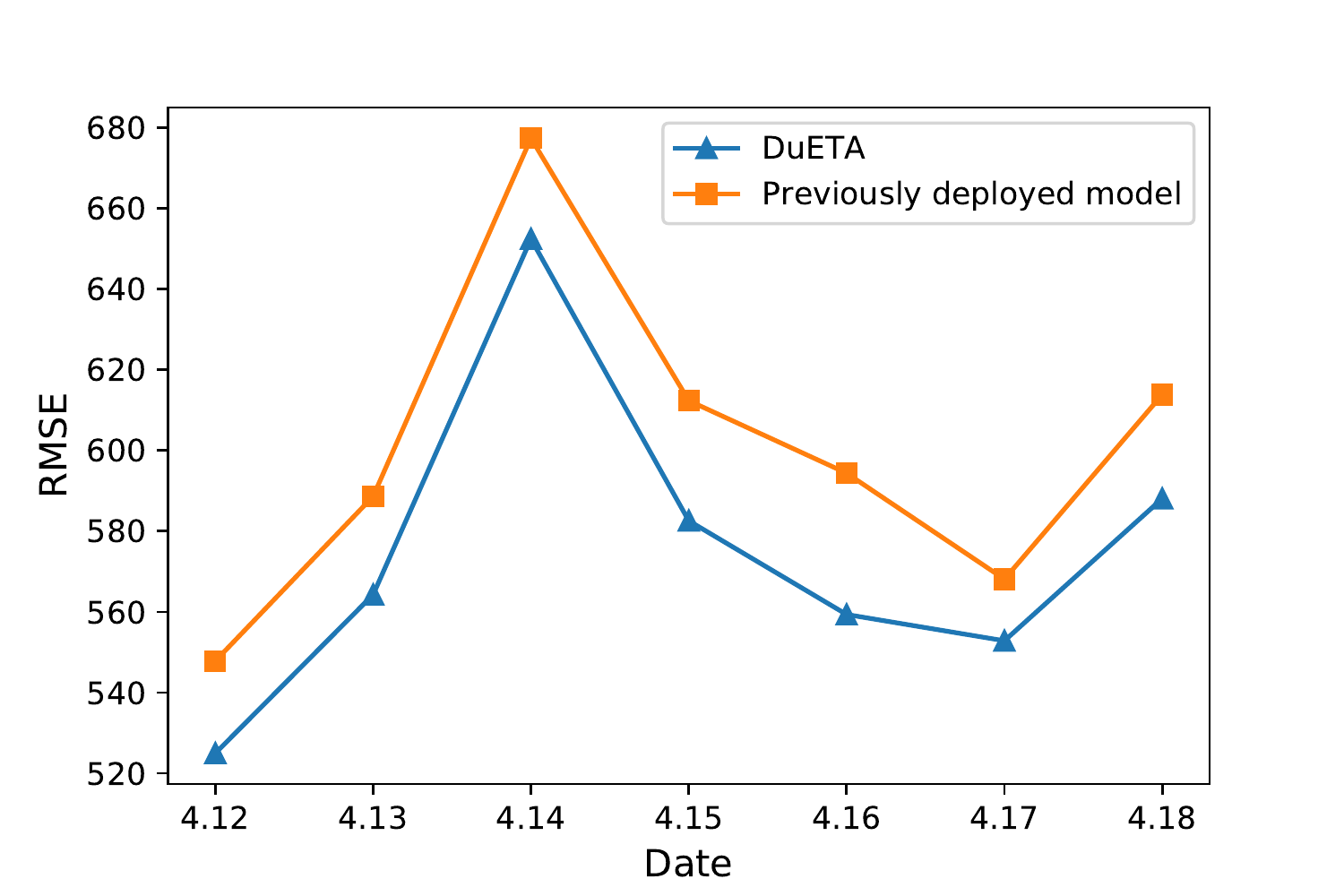}
		\caption{Non-rush hours.}
		\label{fig:online_case4}
	\end{subfigure}
	\begin{subfigure}{0.33\linewidth}
	    \setlength{\abovecaptionskip}{0.1cm}
        \setlength{\belowcaptionskip}{0.01cm}
		\centering
		\includegraphics[width=1.0\columnwidth,trim={0.0cm 0.0cm 0.0cm 0.0cm},clip]{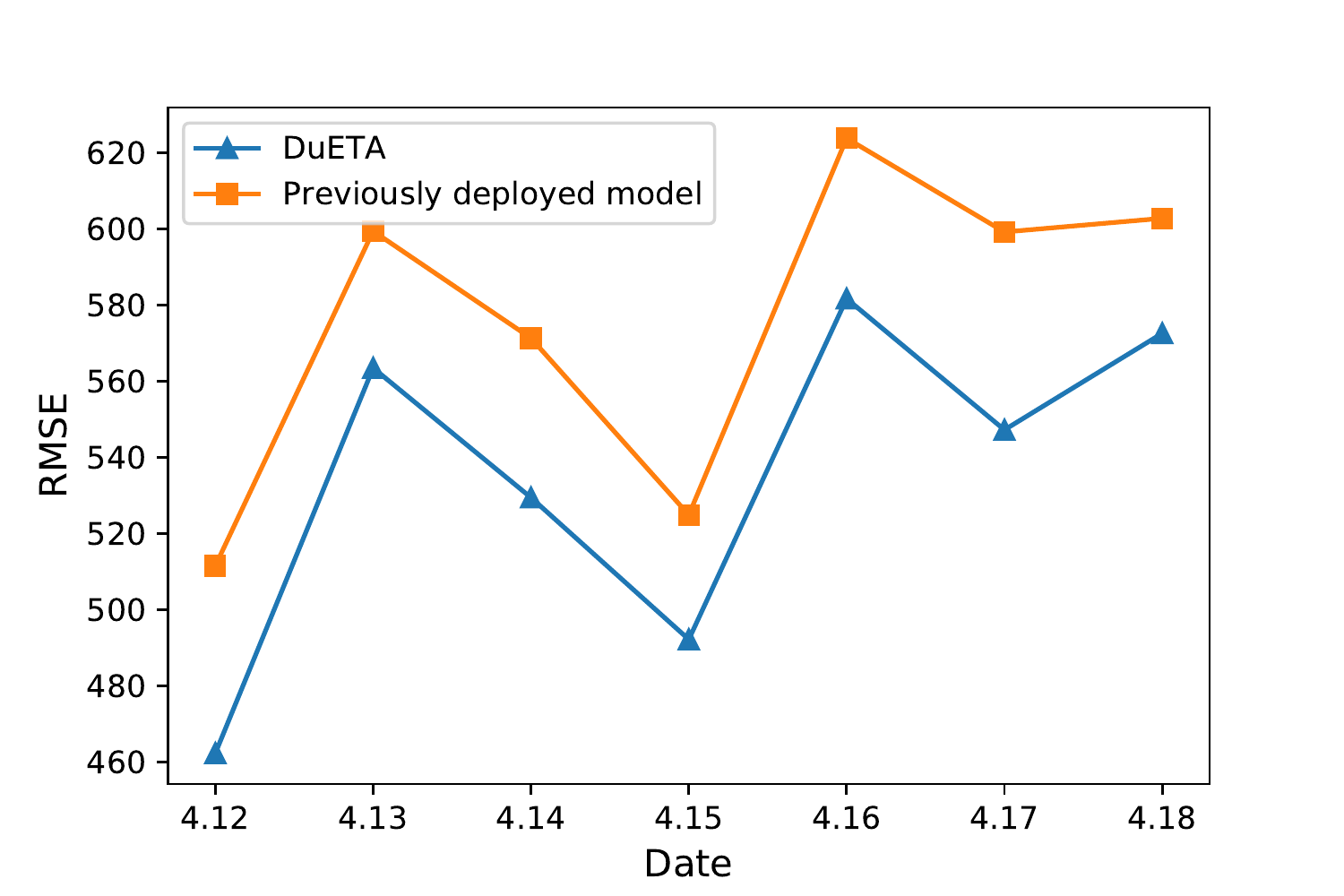}
		\caption{Morning rush hours (7:00-9:00AM).}
		\label{fig:online_case5}
	\end{subfigure}
		\begin{subfigure}{0.33\linewidth}
	    \setlength{\abovecaptionskip}{0.1cm}
        \setlength{\belowcaptionskip}{0.01cm}
		\centering
		\includegraphics[width=1.0\columnwidth,trim={0.0cm 0.0cm 0.0cm 0.0cm},clip]{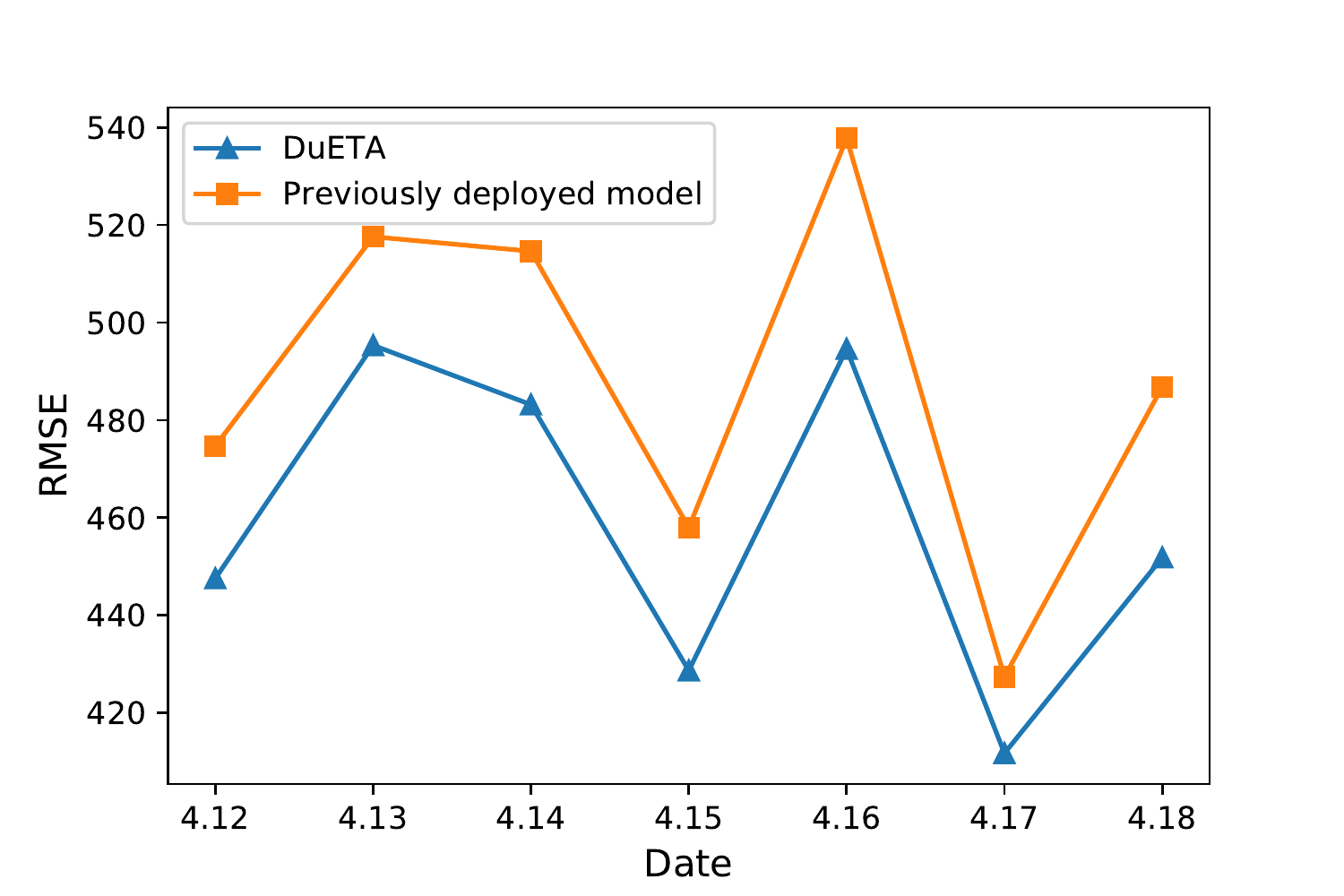}
		\caption{Evening rush hours (5:00-7:00PM).}
		\label{fig:online_case6}
	\end{subfigure}
\caption{Online evaluation of DuETA and the previously deployed model during April 12th - 18th, 2022, in Beijing.}
\label{fig:online_cases}
\vspace{-4mm}
\end{figure*}

Before being launched in production, we would routinely deploy the new ETA prediction model online and make it randomly serve about 25\% of ETA prediction requests. During the period of A/B testing, we monitored the performance of DuETA and compare it with that of the previously deployed model online.
This period conventionally lasted for one week, from Apr. 12th to Apr. 18th, 2022 in Beijing, China.
Figure \ref{fig:online_cases} shows the experimental results.
From the results, we have the following observations.

First, for overall performance comparison, the RMSE scores of DuETA (blue line) are lower than those of the previously deployed model (orange line) from Figure~\ref{fig:online_case1}.
This demonstrates the superiority of DuETA over the previously deployed model.

Second, we further investigate the contribution of DuETA to the travel time estimations of the long travel routes ($\ge$3km) and short travel routes ($<$3km). By comparing Figure~\ref{fig:online_case2} and Figure~\ref{fig:online_case3}, we can observe that DuETA achieves greater improvement on the long travel routes ($\ge$3km) than on the short travel routes ($<$3km). This observation is in line with our expectations, since DuETA focuses on resolving the issue of traffic congestion propagation, especially the cumulative effect of long-distance congestion propagation.
 
Third, we analyze the evaluation results of DuETA on the ETA prediction requests in non-rush hours, morning rush hours, and evening rush hours, as shown in Figure~\ref{fig:online_case4}, Figure~\ref{fig:online_case5}, and Figure~\ref{fig:online_case6}. DuETA consistently surpasses the previously deployed model in all the scenarios across the whole week. Especially since DuETA can respond quickly to the changes in the real-time traffic conditions, the improvement is more significant in the morning and evening rush hours.

Fourth, we observe that the averaged RMSE scores in the online evaluation of DuETA are higher than those in the offline evaluation.
The main reason is that the real-time traffic data processing procedures (e.g., data collection, data identification, and data cleansing) in online settings typically introduce more noise data, which inevitably leads to inflated variance of data distribution between online and offline evaluations.
This results in relatively higher RMSE scores in the online evaluation.
Similar performance gaps between online and offline evaluations are also reported in our previous studies \cite{fang2020constgat,huang2020poiac} at Baidu Maps.

\section{Related Work}
Here we briefly review closely related work in the fields of estimated time of arrival (ETA) prediction and traffic prediction with graph neural networks (GNNs).

\subsection{ETA Prediction}
The mainstream methods for ETA prediction can be categorized into two groups: segment-based methods and end-to-end methods. The segment-based methods \cite{amirian2016predictive,wang2019simple,wang2014travel} are widely used in most navigation services. They first estimate the travel time of each road segment independently. Then, the travel time of the entire route is obtained by simply summing up the estimated travel times of all road segments in the route. The segment-based methods are computationally efficient and scalable, since the travel times of road segments can be estimated in parallel. Although they are efficient, they do not account for the information of the travel route. The route information, such as the connections of road segments and traffic lights, is necessary for ETA prediction.

By contrast, the end-to-end methods \cite{wang2018will,wang2018learning,zhang2018deeptravel,fang2020constgat} take a route as input and directly estimate the travel time of the entire route. 
Compared with the segment-based methods, the end-to-end methods have achieved further improvements because they have taken into account the contextual information of a route  \cite{fang2020constgat}.
For example, \citet{wang2018will} applied convolution on traveling sequence for spatial representation learning and stacked LSTM for temporal modeling. 
\citet{wang2018learning} proposed a Wide-Deep-Recurrent model using a Wide\&Deep network for feature extraction and an LSTM for trajectory information. 
However, the step-by-step message-passing techniques used by most existing methods are inefficient for modeling the traffic congestion propagation patterns along the route. 
In this work, we propose a more efficient method to model traffic congestion propagation patterns, which has been shown to significantly improve ETA prediction performance.

\subsection{Traffic Prediction with GNNs}
GNNs \cite{kipf2016semi,velivckovic2017graph} have been proven to be powerful structural modeling approaches. 
Recent studies have proposed variations of different spatial-temporal GNNs (STGNNs) to tackle the tasks of traffic prediction \cite{dutraffic2022,guo2019attention,he2018stann,li2018diffusion,yu20193d,jiang2021graph,9352246,lee2022learning} and ETA prediction \cite{fang2020constgat,didi2020heta,google2021eta}.
STGNNs usually process spatial-temporal signals using a graph convolution network (GCN) \cite{kipf2016semi} for geographic information and a recurrent model for a temporal dynamic. 
The major drawback of GNNs is their relatively weak scalability on real-world industrial datasets, since increasing the depth of a GNN often means exponential expansion of the neighbor scope. 
To alleviate this problem, recent studies of GNNs \cite{ying2018graph,zeng2021decoupling} suggested that a properly extracted subgraph, consisting of a small number of critical neighbors while excluding irrelevant ones,  can achieve significant accuracy improvement with orders of magnitude reduction in computation and hardware cost.
Inspired by this observation, we propose a congestion-sensitive graph for traffic prediction in this paper.

\section{Conclusions}
Traffic congestion propagation pattern modeling is of great importance for ETA prediction. 
To address this, we develop a novel and practical ETA framework named DuETA.
DuETA can efficiently learn the traffic propagation patterns through an elaborately designed, congestion-sensitive graph and a route-aware graph transformer. 
Experiments show that DuETA is a practical and robust solution for large-scale ETA prediction services.

In the future, we consider addressing the following open problems. 
First, given the observation that roads are successively constructed and upgraded \cite{duare2022,duarus2022}, we plan to investigate the transferability of our model to deal with unseen road segments or regions.
Second, given the observation that the travel times of some routes have a considerable correlation with the POIs distributed along the roads. For example, the roads that pass schools, hospitals, and markets tend to be congested at specific times, which could potentially impact the ETA prediction performance. To address this issue, we plan to utilize the POI retrieval system \cite{huang2020poiac,p4ac-kdd21,hgamnkdd21} as an auxiliary tool to forecast which POIs would be densely populated and how extensively they would affect the ETA prediction.

\clearpage

\bibliographystyle{ACM-Reference-Format}
\balance
\bibliography{main}

\end{document}